\documentclass[letterpaper]{article} 
\usepackage{aaai25}  
\usepackage{times}  
\usepackage{helvet}  
\usepackage{courier}  
\usepackage[hyphens]{url}  
\usepackage{graphicx} 
\usepackage{natbib}  
\usepackage{caption} 
\usepackage{algorithm}
\usepackage{algorithmic}

\usepackage{newfloat}
\usepackage{listings}
\DeclareCaptionStyle{ruled}{labelfont=normalfont,labelsep=colon,strut=off} 
\floatstyle{ruled}
\newfloat{listing}{tb}{lst}{}
\floatname{listing}{Listing}
\usepackage{color}
\usepackage{bbding}
\usepackage{multirow}

\title{ALLVB: All-in-One Long Video Understanding Benchmark}
\author{
    Xichen Tan\textsuperscript{\rm 1}, Yuanjing Luo\textsuperscript{\rm 1}, Yunfan Ye\textsuperscript{\rm 2}, Fang Liu\textsuperscript{\rm 2}\thanks{Corresponding author.}, 
    Zhiping Cai\textsuperscript{\rm 1}\\
}
\affiliations{
    \textsuperscript{\rm 1}College of Computer Science and Technology, National University of Defense Technology, Changsha, China\\
    \textsuperscript{\rm 2}School of Design, Hunan University, Changsha, China\\
    tanxc23@nudt.edu.cn

}

\usepackage{bibentry}

\begin{document}

\maketitle

\begin{abstract}
From image to video understanding, the capabilities of Multi-modal LLMs (MLLMs) are increasingly powerful. However, most existing video understanding benchmarks are relatively short, which makes them inadequate for effectively evaluating the long-sequence modeling capabilities of MLLMs. This highlights the urgent need for a comprehensive and integrated long video understanding benchmark to assess the ability of MLLMs thoroughly. To this end, we propose ALLVB (ALL-in-One Long Video Understanding Benchmark). ALLVB's main contributions include: 1) It integrates 9 major video understanding tasks. These tasks are converted into video QA formats, allowing a single benchmark to evaluate 9 different video understanding capabilities of MLLMs, highlighting the versatility, comprehensiveness, and challenging nature of ALLVB. 2) A fully automated annotation pipeline using GPT-4o is designed, requiring only human quality control, which facilitates the maintenance and expansion of the benchmark. 3) It contains 1,376 videos across 16 categories, averaging nearly 2 hours each, with a total of 252k QAs. To the best of our knowledge, it is the largest long video understanding benchmark in terms of the number of videos, average duration, and number of QAs. We have tested various mainstream MLLMs on ALLVB, and the results indicate that even the most advanced commercial models have significant room for improvement. This reflects the benchmark's challenging nature and demonstrates the substantial potential for development in long video understanding.
\end{abstract}

%
\begin{links}
    \link{Datasets}{https://huggingface.co/datasets/ALLVB/ALLVB}
\end{links}

\begin{figure*}[ht]
    \begin{center}
    \centerline{\includegraphics[width=1\linewidth]{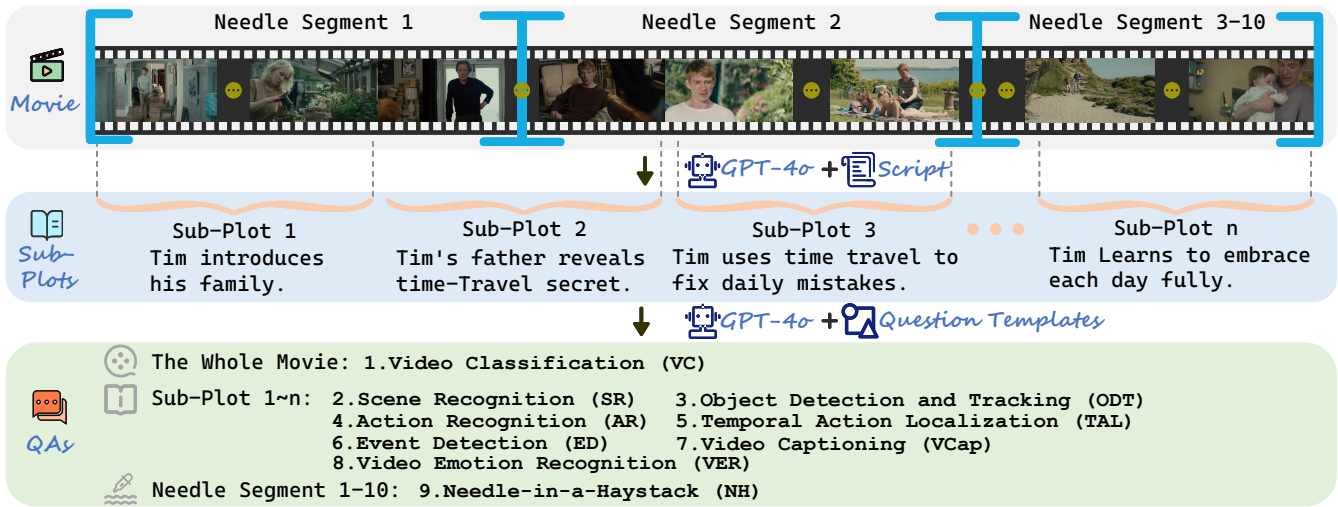}}
         \caption{The construction pipeline of ALLVB. Utilizing the powerful processing capabilities of GPT-4o, we first segment the movie into different sub-plots based on the corresponding script content. We then create Q\&As for the entire video, each sub-plot, and evenly divided needle segments using 91 question templates. Note that needle segments do not correspond to the sub-plot segments.}
     \label{pipeline}
     \end{center}
\end{figure*}

\section{Introduction}
The field of Large Language Models (LLMs) is currently a highly popular research area, rapidly evolving from text-focused to multi-modal, incorporating image and video inputs. To objectively assess the performance of these models, various Q\&A benchmarks are continually being proposed.

For pure text benchmarks, notable examples include MMLU~\cite{hendrycks2020measuring}, which evaluates LLMs across 57 tasks in various academic fields, and AGIEval~\cite{zhong2023agieval}, which focuses on performance in standardized exams like GRE, GMAT, and China's Gaokao. For image-text multi-modal benchmarks, ScienceQA~\cite{lu2022learn} includes Q\&As from elementary and middle school curricula, while MMMU~\cite{yue2024mmmu} features questions from six core academic subjects at the university level. 

Regarding video-text benchmarks, current benchmarks mainly focused on short videos, such as MSVD-QA~\cite{xu2017video}, MSRVTT-QA~\cite{xu2017video}, TGIF-QA~\cite{jang2017tgif}, ActivityNet-QA~\cite{yu2019activitynet}, and BDIQA~\cite{mao2024bdiqa}. These benchmarks typically involve videos averaging under 10 seconds, except for ActivityNet-QA, which averages 180 seconds.

When it comes to benchmarks for long videos, several contemporary works are noteworthy, including MLVU~\cite{zhou2024mlvu}, Video-MME~\cite{fu2024video}, and LVBench~\cite{wang2024lvbench}. They rely on manual Q\&As annotations, limiting scalability due to labor costs, and resulting in shorter videos and fewer Q\&As per video. Moreover, the lack of a unified standard in designing video comprehension questions poses challenges. To address these issues, we develop an automated pipeline for a more efficient and scientifically robust benchmark.

Since existing LLMs do not yet natively support video modality input, the current method involves generating textual descriptions of video content for the models. Given the significant challenges of describing hour-long videos, we have opted to use existing text descriptions closely related to the video content, with movie scripts being the most common example.

First, we collect a large number of movie scripts from open-source websites, then filter out duplicates and unsuitable scripts, ultimately obtaining 1,376 movie scripts. These scripts are used as input to GPT-4o~\cite{openai2024gpt4o} for generating video-related Q\&As. To ensure that the LLM can capture the details within the movie and considering its limitations in handling ultra-long contexts, we use a two-stage segmentation method to divide the script into different plots and sub-plots, and then construct overarching Q\&As for the entire script and detailed Q\&As for the sub-plots. This method ensures the correctness of the Q\&As construction.

To design the Q\&As as objectively as possible and to enhance the benchmark's versatility, we select 9 major existing video understanding tasks and expand them into 91 sub-tasks, designing corresponding question templates for each. These question templates comprehensively assess the MLLMs' abilities in summarization, information extraction, temporal reasoning, and more. GPT-4o then generates the final Q\&As using these 91 templates and the video content. 

To ensure the correctness of the generated Q\&As, we implement strict quality control measures for ALLVB. First, during the plot segmentation phase, scripts are used to verify the continuity and completeness of the generated plots. In the Q\&As construction phase, scripts ensure that the format of questions and answers complies with the designed rules. Finally, the paper's authors and recruited volunteers conduct a three-stage manual review, with details provided in the quality control section below.

The above provides an overview of the automated pipeline used to construct ALLVB. This pipeline distinguishes ALLVB from existing long video understanding benchmarks in several key aspects:

1. We design 91 question templates for 9 types of video understanding tasks, effectively integrating existing tasks into a comprehensive video Q\&A framework. This approach thoroughly evaluates the MLLMs' ability to understand long videos, providing substantial practical value.

2. The benchmark leverages external text information and GPT-4o's powerful processing capabilities, utilizing a custom-designed, fully automated pipeline with rigorous quality control. This ensures its correctness, excellent scalability, and simplifies future maintenance.

3. ALLVB includes 1,376 long videos, averaging nearly two hours each, with a total of 252k Q\&As, or 183 Q\&As per video. In the field of long video understanding benchmarks, its average length is nearly \(2 \times\) longer than the second longest benchmark, and its total number of Q\&As is \(11.5 \times\) greater than the second largest benchmark. From these two dimensions, ALLVB is the most comprehensive and largest long video understanding benchmark to date.

\section{Related Work}
\subsection{Multi-model LLMs}

Following the introduction of ChatGPT~\cite{openai2024chatgpt}, considerable effort~\cite{darec} has been made to combine vision encoders with pre-trained LLMs to create MLLMs, enabling support for both textual and visual inputs. Image-based MLLMs such as Otter-I~\cite{li2023otter} and LLaVA-1.6~\cite{liu2024llava} use MPT~\cite{mosaicml2023introducing} and Vicuna~\cite{chiang2023vicuna} as base models, respectively, and incorporate CLIP's ViT-L/14~\cite{radford2021learning} for processing multiple image inputs. Video-based MLLMs, often built on Vicuna or LLaMA~\cite{touvron2023llama}, also utilize ViT~\cite{dosovitskiy2020image} encoders to handle video data, where each adopts different frame processing techniques:  mPLUG-Owl-V~\cite{ye2023mplug} uses a set of learnable tokens to summarize visual information, MovieChat~\cite{song2024moviechat} employs methods from ToMe~\cite{bolya2022token} to merge similar tokens between adjacent frames, and LLaMA-VID~\cite{li2023llama} directly applies average pooling to reduce the number of image tokens.

Despite progress, MLLMs face challenges in long video comprehension, requiring expanded inferential capabilities. Models like MovieChat and LLaMA-VID, meant for long videos, have yet to demonstrate proven performance on hour-long videos. A reliable long video benchmark is needed for objective and accurate evaluation of MLLMs.

\subsection{Video Understanding Benchmarks}
Video understanding benchmarks assess LLMs' capabilities in video analysis through a range of tasks. Most, like MSVD-QA, MSRVTT-QA, TGIF-QA, and ActivityNet-QA, concentrate on short videos. They transform video captions into Q\&As (MSVD-QA, MSRVTT-QA, and TGIF-QA) or derive Q\&As from the Video Classification task (ActivityNet-QA), often using crowdsourcing, which is labor-intensive. Similarly, MovieQA~\cite{tapaswi2016movieqa} is another movie-based short-form video benchmark that, like the others, is manually annotated and has an average length of just 203 seconds.

For long video understanding, there are some benchmarks such as MLVU, Video-MME, LVBench, and MoVQA~\cite{zhang2023movqa} (which also uses movies), that have manually annotated through various mechanisms: MLVU categorizes Q\&As into three types: holistic LVU, single-detail LVU, and multi-detail LVU. Video-MME constructs Q\&As from the perspectives of perception, reasoning, and information synopsis. LVBench builds Q\&As based on six core capabilities: Temporal Grounding, Summarization, Reasoning, Entity Recognition, Event Understanding, and Key Information Retrieval. MoVQA classifies all Q\&As into six types: Information Synopsis, Temporal Perception, Spatial Perception, Causal Reasoning, Hypothetical Reasoning, and External Knowledge.

We believe these approaches to constructing Q\&As are reasonable, but they highlight a lack of a unified standard for constructing Q\&As, and there is room for improvement in terms of video duration and size. We notice that these approaches closely resemble existing video understanding tasks. To avoid subjectively classifying Q\&As, our approach, using established video understanding tasks to construct question templates and generating Q\&As with GPT-4o, aims to create more objective and aligned Q\&As, as compared in detail in Tab.~\ref{comparisons}.

\section{ALLVB: All-in-One Long Video Understanding Benchmark}
In this section, we will provide a detailed explanation of the benchmark construction pipeline and quality control measures and present the specific statistics of ALLVB. We will also compare these with existing benchmarks to highlight ALLVB's distinctiveness.

\begin{figure*}[ht]
    \begin{center}
    \centerline{\includegraphics[width=1\linewidth]{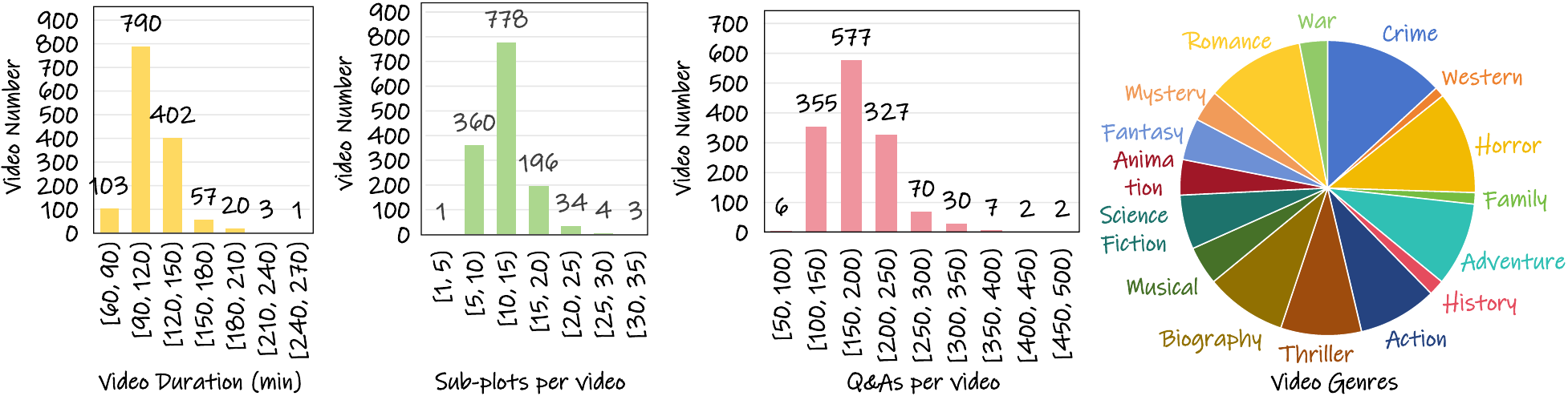}}
     \caption{ALLVB Benchmark Statistics Chart. The distributions shown, from left to right, are \textbf{video duration}, \textbf{number of sub-plots} per video, \textbf{number of Q\&As} per video, and \textbf{video genres}. Most videos are between 90-150 minutes in length, which is significantly longer than those in other benchmarks, highlighting the challenge of ALLVB. The majority of videos are divided into 5-20 sub-plots, resulting in most videos having 100-250 Q\&As, showcasing the benchmark's comprehensiveness. Finally, our videos span 16 diverse genres, ensuring the benchmark's general applicability.}
     \label{benchmark stats}
     \end{center}
\end{figure*}

\begin{table*}[t]
    \begin{center}
\resizebox{\linewidth}{!}
{
\begin{tabular}{c|c|c|c|c|c|c|c}
\hline
Benchmarks & \#Videos & Avg.Len.(min) & \#Q\&As & \# Avg. Q\&As & Q\&As type & Anno. & Subs. \\ \hline
\multicolumn{8}{c}{\textcolor[RGB]{128,128,128}{\textit{Short Video Understanding Benchmarks}}} \\
MovieQA~\cite{tapaswi2016movieqa} & 408 & 3.38 & 14,944 &   36& MC & M & \Checkmark\\
MSVD-QA~\cite{xu2017video}& 1,970 & 0.16  &  50,505 & 25 & OE & A & \XSolidBrush \\
MSRVTT-QA~\cite{xu2017video}& 10,000 & 0.25 & 243,680 & 24 & OE & A & \XSolidBrush \\
TGIF-QA~\cite{jang2017tgif} & 71,741 & 0.05 &  165,165  & 2 & OE\&MC & A\&M &  \XSolidBrush \\
TVQA~\cite{lei2018tvqa} & 21,793 & 1.26 & 152,545 & 7 & MC & M & \Checkmark \\
ActivityNet-QA~\cite{yu2019activitynet} & 5,800 & 3 & 58,000 & 10 & OE & M & \XSolidBrush \\
How2QA~\cite{li2020hero} & 9,035 & 1 & 44,007 & 5 & MC & M & \XSolidBrush \\
NExT-QA~\cite{xiao2021next} & 5,440 & 0.73 & 52,044 & 9 & OE\&MC & M & \XSolidBrush\\
MVBench~\cite{li2024mvbench} & 3,641 & 0.26 & 4,000 & 1 & MC & A & \XSolidBrush \\
CinePile~\cite{rawal2024cinepile} & 9,396 & 2.66 & 303,828 & 32 & MC & A\&M & \Checkmark \\
EgoSchema~\cite{mangalam2024egoschema} & 5,063 & 3 & 5,063 & 1  & MC & A\&M & \XSolidBrush\\ \hline
\multicolumn{8}{c}{\textcolor[RGB]{128,128,128}{\textit{Long Video Understanding Benchmarks}}} \\
MoVQA~\cite{zhang2023movqa} & 100 &  16.53 &  21,953 &  \textbf{219} & MC &  M & \Checkmark\\
MLVU~\cite{zhou2024mlvu} & 1,334 & 12 & 2,593 & 2& OE\&MC & M & \XSolidBrush \\  
Video-MME~\cite{fu2024video}& 900 & 17 & 2,700  & 3 & MC & M &\Checkmark \\ 
LVBench~\cite{wang2024lvbench} & 103 &  68 & 1,549 & 15  & MC & M & \XSolidBrush \\ \hline
\textbf{ALLVB}& \textbf{1,376} & \textbf{114.62} & \textbf{252,420} & 183 &  MC & A & \Checkmark\\ \hline
\end{tabular}
}
    \end{center}
    \caption{Comparison with other benchmarks, where the abbreviations are defined as follows: \textbf{Avg.Len.} (Average length of each video), \textbf{Avg. Q\&As} (Average number of Q\&As per video), \textbf{OE} (Open-Ended questions), \textbf{MC} (Multiple-Choice questions), \textbf{Anno.} (Annotation Method), \textbf{A} (Automatic Annotation), \textbf{M} (Manual Annotation), \textbf{Subs.} (Subtitles). In the realm of long video benchmarks, ALLVB leads in terms of the number of videos, average video length, and the quantity of Q\&As.}
    \label{comparisons}
\end{table*}

\subsection{Benchmark Construction Pipeline}
\paragraph{Data Collection and Cleaning.}
We initially scrape 2,520 scripts from Script Reader Pro\footnote{https://www.scriptreaderpro.com} and SWN\footnote{https://www.screenwritersnetwork.org/script/?page=1}. Through regularized detection of script names (e.g., names containing ``episode'' or season numbers indicating TV scripts), we identify 711 TV scripts and 1,809 movie scripts. Given that TV episodes are shorter in duration and often have plots spread across multiple episodes, we decide to temporarily exclude TV scripts from annotation.

For the remaining 1,809 movie scripts in PDF format, we first filter out abnormal files with very small sizes (\(<\)10KB), as these typically do not contain complete script content. We then use a script to extract text from PDFs. Following this, we further filter out scripts with too few text characters (\(<\)3,000 characters) and those containing an excessive number of non-text characters, such as some early movie scripts in blurry PDF formats. After these steps, we ultimately obtain 1,376 text-formatted movie scripts that meet our criteria.

Finally, we download 1,376 corresponding movies from YTS\footnote{https://ww4.yts.nz/} and their respective subtitles from YTS and SUBDL\footnote{https://subdl.com/}.

\paragraph{Data Preprocessing: Plot Segmentation.}
Previous long video annotations involve humans watching the video and asking questions based on the content according to specified requirements. This approach limits the number of Q\&As per video and often overlooks numerous details in long videos. To ensure that the Q\&As generated by GPT-4o more thoroughly cover the details in the video, we instruct GPT-4o to first segment the movie script into different continuous plots based on its content. During this process, we discover that single-pass segmentation still results in plots that span considerable durations, so we adopt a two-stage segmentation strategy. In this strategy, sub-plots are further divided within the initially segmented plots, enabling finer-grained video segmentation and ensuring the correctness of the generated questions.

\paragraph{Question Template Design.}
The tasks addressed by the source videos of short video benchmarks and the tasks considered when constructing Q\&As for long video benchmarks are very similar to existing video understanding tasks. Therefore, we design our question templates directly based on these established video understanding tasks. We have collected 9 common types of video understanding tasks suitable for conversion into Q\&As, and these 9 tasks comprehensively cover the content assessed by other short video or long video understanding benchmarks, such as reasoning, summarization, recognition, and more. The 9 tasks include: \textit{Video Classification (VC), Scene Recognition (SR), Object Detection and Tracking (ODT), Action Recognition (AR), Temporal Action Localization (TAL), Event Detection (ED), Video Captioning (VCap), Video Emotion Recognition (VER), and Needle-in-a-Haystack (NH) task for video}. For the 9 tasks, we have GPT-4o provide diverse and generalized sub-tasks. From these, we select 11, 12, 10, 11, 7, 12, 8, 10, and 10 non-overlapping sub-tasks that comprehensively assess various capabilities of MLLMs, further expanding the scope of the original tasks

For the ``Needle-in-a-Haystack" (NH) task, a common approach is to insert content unrelated to the original video and then ask questions about it. For example, Gemini 1.5 Pro~\cite{reid2024gemini} overlays the text "The secret word is needle" on a single randomly sampled video frame and then asks the LLM, "What is the secret word?" However, advanced commercial models might recognize these as tests and refuse to answer. To address this, we ask detailed questions about the content of specific video frames, thereby avoiding this issue and preventing redundancy with the other 8 task types, such approach is more challenging and can provide a better assessment of the LLM's retrieval capabilities in long contexts. Every video is evenly divided into 10 segments, with each segment containing one sub-task that targets specific details in the frames. 

Finally, we design 91 corresponding question templates based on these 91 sub-tasks.

\paragraph{Constructing Q\&As.}
\textit{1) Number of Q\&As.} First, it is important to note that the Video Classification task applies to the entire video, while the Needle-in-a-Haystack task is applied to each evenly divided video segment. The remaining 7 tasks generate questions for each sub-plot of the video. To avoid excessive homogeneous questions in the benchmark, we randomly select 2 question templates from each of the 7 tasks to design questions for each sub-plot. Assuming a movie has \(n\) sub-plots, the corresponding number of Q\&As would be: \(11 (VC) + 2*7*n + 10 (NH) = 21 + 14n\).

\textit{2) Questions.} For the Video Classification task, we input the entire script content and question templates into GPT-4o, prompting it to design Q\&As based on the requirements outlined in the prompt. For questions targeting sub-plots, we instruct GPT-4o to first describe the scene content of the sub-plot in the question, helping to locate the question within the video and reduce ambiguity. For Needle-in-a-Haystack questions, we evenly divide the video into segments and sample 11 continuous frames from a random position in each segment at a frame rate of 1 fps. GPT-4o then describes the scene corresponding to these 11 frames and generates detailed questions about the middle frame or a specific frame.

\textit{3) Answers.} All Q\&As are presented as multiple-choice questions, with each question offering 5 options, one correct answer, and 4 distractors. This format allows for easy calculation of accuracy during testing and eliminates the subjective judgment issues associated with open-ended answers. 

To ensure the scientific validity of the options, we require that all options be of similar length. The incorrect options must be related to the video content, and the correct option should be randomly distributed among the incorrect ones. We also stipulate that the answers cannot be deduced simply by examining the question content alone. Additionally, GPT-4o is required to provide reasoning when generating the answers to support the deduction process and enhance the correctness of the output.

\paragraph{Quality Control.}
\textit{During the plot segmentation phase}, we use regularization methods to verify that the start and end positions of the main plots and sub-plots align correctly, ensure the continuity of the sub-plots, and confirm that the sub-plots comprehensively cover all the script content. Additionally, dividing the script content into different sub-plots before designing questions can ensure the correctness of the generated Q\&As.

\textit{During the Q\&As construction phase}, we also use regularization methods to determine whether all the questions are generated completely and whether there are any errors in the question content. For instance, if a question mentions the movie title, it could lead to information leakage. If a question related to a sub-plot lacks specific scene descriptions, or if a Needle-in-a-Haystack question references the frame number, these are considered invalid. We regenerate any questions that do not meet the requirements and conduct secondary verification. We also analyze the distribution of all the correct options and use a script to randomly adjust them. As a result, the proportion of correct options being A, B, C, D, and E is 18\%, 20\%, 21\%, 19\%, and 22\%, respectively, which is close to an even distribution.

\textit{Manual reviewing}. Although movie scripts generally align with the movie content, discrepancies can occur in certain details, which may affect the correctness of the answers. To address this issue, all the authors of this paper, along with 12 recruited volunteers, conduct a 3-stage manual review. \textbf{The first stage} focuses on identifying potentially problematic question types. We find that tasks related to video classification, scenes, events, and emotion analysis are almost entirely correct, as the script and movie content typically match. The issues primarily arise with object and action recognition tasks, such as identifying the color of an object, counting objects, or recognizing actions. \textbf{In the second stage,} we filter these questions from the benchmark and manually verify them against the movie content. For questions with discrepancies, we make corrections or regenerate them to ensure the correctness of the answers. \textbf{In the third stage}, we conduct another round of manual review to ensure that no errors are found. These rigorous quality control steps ultimately ensure the high quality of ALLVB. An example of using the pipeline to fully construct Q\&As from a single video is shown in Fig.~\ref{pipeline}.

\subsection{Benchmark Statistics}

Through the GPT-4o automated benchmark construction pipeline, we collect a total of 1,376 high-quality movies and segment them into 15,966 sub-plots based on the script content, averaging 11.6 sub-plots per movie. Using these segmented sub-plots and entire movies, GPT-4o generates 252,420 Q\&As according to 91 question templates, averaging 183 questions per movie. For more detailed data, please refer to Fig.~\ref{benchmark stats}. We also provide a detailed comparison with existing video understanding benchmarks in Tab.~\ref{comparisons}, and the following points are worth noting:
\begin{itemize}
\item In the realm of long videos, ALLVB features the highest number of videos, the longest average duration, and the most Q\&As. This is made possible by our automated pipeline, which facilitates benchmark expansion and maintenance, significantly reducing labor costs.
\item We select multiple-choice questions as the format to enable objective evaluation. Additionally, we provide subtitles for each video, allowing for questions that do not rely solely on video content. Many benchmarks avoid this issue by describing characters' physical features instead, which does not fully assess the model's ability to extract information from multiple inputs.
\end{itemize}

\begin{table*}[t]
    \begin{center}
\resizebox{\linewidth}{!}
{
\begin{tabular}{c|c|c|c|c|c|c|c|c|c|c|c}
\hline
\multirow{2}{*}{Models}& \multirow{2}{*}{LLM}  & \multicolumn{9}{c|}{Video Understanding Tasks} & \multirow{2}{*}{Avg. Acc. (\%)}\\ \cline{3-11}
& & VC & SR & ODT & AR & TAL & ED & VCap & VER & NH & \\ \hline
\multicolumn{12}{c}{\textcolor[RGB]{128,128,128}{\textit{Image MLLMs}}} \\ 
Otter-I~\cite{li2023otter} & MPT-7B  & 37.7 & 30.1 & 25.1 & 22.8 & 22.8 & 26.4 & 29.1 & 31.7 & 19.6 & 26.9 \\
LLaVA-1.6~\cite{liu2024llava} & Vicunna-7B  & 68.4 & 51.3 & 39.0 & 43.6 & 43.3 & 46.1 & 56.2 & 57.3 & 32.3 & 48.5 \\
GPT4-Turbo~\cite{achiam2023gpt} & GPT4 & 84.5 & 59.7 & 51.9 &58.5 & 53.5 & 63.2 & 72.2 & 67.5 & 32.9 & 60.8 \\ \hline
\multicolumn{12}{c}{\textcolor[RGB]{128,128,128}{\textit{Video MLLMs}}} \\ 
Otter-V~\cite{li2023otter} & LLaMA-7B  & 25.6 & 23.6 & 22.3 & 22.8 & 25.9 & 24.4 & 22.7 & 20.0 & 19.6 & 23.1 \\
mPlug-Owl-V~\cite{ye2023mplug} & LLaMA-7B & 22.8 & 25.6 & 21.5 & 25.4 & 22.0 & 25.2 & 23.8 & 22.3 & 17.8 & 23.3 \\
MovieChat~\cite{song2024moviechat} & Vicunna-7B & 26.2 & 25.6 & 23.0 & 23.8 & 23.0 & 26.7 & 25.1 & 24.6 & 20.9 & 24.4 \\
VideoChat~\cite{li2023videochat} & Vicunna-7B & 35.2 & 31.8 & 26.3 & 29.4 & 29.5 & 30.9 & 31.6 & 34.3 & 23.2 & 30.4 \\
VideoChat2~\cite{li2024mvbench} & Vicunna-7B  & 52.3 & 34.3 & 31.9 & 36.9 & 36.0 & 37.0 & 38.1 & 44.7 & 34.6 & 37.8 \\
LLaMA-VID~\cite{li2023llama} & Vicunna-7B & 71.4 & 45.5 & 41.3 & 39.8 & 35.8 & 44.8 & 57.4 & 56.9 & 27.4 & 46.5 \\
TimeChat~\cite{ren2024timechat} & LLaMA-2 7B & 61.8 & 52.1 & 41.1 & 46.9 & 38.8 & 48.4 & 53.1 & 49.5 & 29.8 & 47.1 \\
GPT-4o~\cite{openai2024gpt4o} & GPT4  & \underline{91.8} & \underline{67.3} & \underline{56.4} & \underline{67.1} & \underline{63.1} & \underline{70.0} & \underline{78.2} & \underline{75.9} & \underline{37.6} & \underline{68.0} \\ 
Claude3.5 Sonnet~\cite{claude35sonnet2024} & Claude 3.5 & \textbf{92.7} & \textbf{73.5} & \textbf{68.2} & \textbf{75.2} & \textbf{73.5} & \textbf{75.4} & \textbf{84.9} & \textbf{82.1} & \textbf{42.5} & \textbf{75.2} \\ \hline
\end{tabular}
}
    \end{center}
    \caption{The test results of various MLLMs on ALLVB, including the accuracy for 9 types of video understanding tasks: \textbf{VC} (Video Classification), \textbf{SR} (Scene Recognition), \textbf{ODT} (Object Detection and Tracking), \textbf{AR} (Action Recognition), \textbf{TAL} (Temporal Action Localization), \textbf{ED} (Event Detection), \textbf{VCap} (Video Captioning), \textbf{VER} (Video Emotion Recognition), and \textbf{NH} (Needle-in-a-Haystack), as well as the \textbf{Avg. Acc.} (Average Accuracy).}
    \label{experiment1}
\end{table*}

\section{Experiments and Analysis}
In this section, we test various MLLMs on ALLVB and thoroughly demonstrate the challenges of ALLVB through analysis of the experimental results.

\subsection{Implementation Details}
\paragraph{Settings.} 
First, we divide ALLVB into a training set and a test set at a 9:1 ratio, containing 1,236 and 140 videos, respectively. The training set can be used for future pre-training or fine-tuning of other MLLMs. Then, we test various open-source and closed-source MLLMs on the test set. Among them,  Otter-I~\cite{li2023otter}, LLaVA-1.6~\cite{liu2024llava} and GPT4-Turbo~\cite{achiam2023gpt} are image MLLMs that support multiple image inputs. Otter-V~\cite{li2023otter}, mPlug-Owl-V~\cite{ye2023mplug}, LLaMA-VID~\cite{li2023llama}, VideoChat~\cite{li2023videochat}, VideoChat2~\cite{li2024mvbench}, MovieChat~\cite{song2024moviechat}, TimeChat~\cite{ren2024timechat}, GPT-4o~\cite{openai2024gpt4o} and Claude 3.5 Sonnet~\cite{claude35sonnet2024} are video MLLMs. Open-source models are run locally on an NVIDIA 4090, while closed-source models are accessed via official API.

To ensure fairness in testing and that each model can perform inference, all models receive 16 frames uniformly sampled across the entire video, along with the corresponding subtitles for these frames. LLM parameters are uniformly set to 7B and all tests are conducted in a 0-shot format. Due to the limitations in handling long contexts, open-source models answer each Q\&A in the video individually, while closed-source models answer all Q\&As in a single video at once. Open-source and closed-source models each use the same prompts to ensure fairness. The specific prompt details are as follows:
\begin{itemize}
\item Prompts for open-source models:
\begin{quote} 
\begin{scriptsize} 
\textbf{System Prompt:}
You are an expert at analyzing videos and their accompanying subtitles. Carefully observe the details in the video frames and their corresponding subtitles. Based on your observations, select the best option for the question provided.

\textbf{User Prompt:}

Frames from a video, their corresponding subtitles, and a multiple-choice question with five options have been provided. Your task is to select the best answer based on the information from the video frames and subtitles.

\quad Subtitles: \{Insert the subtitles here\}

\quad Question: \{Insert the question here\}

Please strictly follow the format below for outputting your answers:

Best Answer: [Correct option]
\end{scriptsize}
\end{quote}

\item Prompts for closed-source models:
\begin{quote}
\begin{scriptsize} 
\textbf{User Prompt:}
Video frames, corresponding subtitles, and \{question\_num\} multiple-choice questions with five options each have been provided. Your task is to select the best answer based on the information from the video frames and subtitles.

\quad Subtitles: \{Insert the subtitles here\}

\quad Question: \{Insert the question here\}

Please strictly follow the format below for outputting your answers:

Question 1: [Correct option]

Question 2: [Correct option]

...

Question \{question\_num\}: [Correct option]
\end{scriptsize}
\end{quote}

\end{itemize}

\paragraph{Evaluation Metrics.}
We extract the answers output by each model and compare them with the correct answers to calculate the final accuracy. Despite strictly specifying the answer format in the prompts, we still encounter a variety of answer formats in the outputs. To address this, we manually create numerous regex-matching scripts tailored for different model outputs. For content without letter options, we calculated the similarity between the output text and the question options to identify the closest answer. We observe that closed-source models require significantly fewer regex rules compared to open-source models. For instance, GPT-4o only needed two rules to identify all output options, while models like mPlug-Owl-V and VideoChat required 34 and 26 regex rules, respectively, to match most output options. This suggests that closed-source models have much stronger instruction-following capabilities compared to open-source models.

\subsection{Results and Analysis}
We calculate the individual accuracy for each model across the  9 tasks, as well as the average accuracy. The detailed results can be found in Tab.~\ref{experiment1}. From this analysis, several key conclusions can be drawn:
\begin{itemize}
    \item \textbf{Claude 3.5 Sonnet achieves the highest performance across all 9 tasks}, with GPT-4o consistently securing the second-best results. Although we are unable to run inference locally with a model that matches Claude's parameter size, these results highlight the superior capabilities of the Claude and GPT series in MLLMs.
    \item \textbf{There is a significant variation in results across different tasks.} Tasks such as VC (Video Classification), VCap (Video Captioning), and VER (Video Emotion Recognition) score noticeably higher. These tasks generally do not require extensive interaction with detailed visuals, different characters, objects, or environments, making them relatively less challenging. On the other hand, tasks like ODT (Object Detection and Tracking), TAL (Temporal Action Localization), and NH (Needle-in-a-Haystack) involve more detailed visual elements and logical interactions, such as object colors, object counts, and action sequences. These tasks are more difficult and test the model's comprehension and reasoning abilities more rigorously. The NH task, in particular, deals with very fine-grained information in the video frames, leading to lower scores across all models.
    \item \textbf{Open-source video MLLMs do not show a significant advantage} over the image MLLMs trained on image data under the same model parameter size (7B) and input conditions. This could be partly due to the limited number of video frames used as input, but it also raises questions about whether most open-source video MLLMs have sufficient advantages in temporal modeling.
    \item \textbf{Even the best closed-source model, Claude 3.5 Sonnet, achieves an accuracy of only 75.2\%.} One reason for this is that, due to the limitations of some open-source models, we standardize the input frame count to 16 to ensure fairness in comparison, which limits the amount of video information the models could access. Additionally, this result indicates that, whether open-source or closed-source, there is still significant room for improvement in retrieving and reasoning within ultra-long video contexts. This also highlights the challenges presented by ALLVB and its potential contributions to the future development of the video MLLMs community.
   
\end{itemize}

\subsection{Further Discussion}
Based on the analysis of the experimental results, we would like to further discuss the following two aspects:

\paragraph{The impact of video length.}
Although the videos in ALLVB are all at the hour-long level, we randomly select three video length distributions ranging from short to long to verify the impact of different video lengths on model accuracy. We then sample four baseline models to analyze their accuracy performance across these varying video lengths. The specific results are presented in Tab~\ref{duration}.

As the video length increases, all models exhibit varying degrees of accuracy decline. Since the input video frames are fixed at 16 frames, longer videos result in more information being missed by the models, making it more challenging to perform effective reasoning and test the models' ability to understand long contexts. We also observe that GPT-4o demonstrates better robustness with longer video durations, as its accuracy remains stable when transitioning from 105-115 minutes to 130-140 minutes. Because the overall video lengths in ALLVB are generally long, the increase in video length does not cause a particularly significant drop in accuracy. However, in general, the performance of multimodal LLMs still tends to decrease as the video length increases.

\begin{table}[t]
    \begin{center}
\resizebox{\linewidth}{!}
{
\begin{tabular}{c|c|c|c}
\hline
\multirow{2}{*}{Models}& \multicolumn{3}{c}{Video Length Distribution (min)} \\ \cline{2-4}
& [80, 90) & [105, 115) & [130, 140) \\ \hline
Otter-V &  24.7 & 24.3 (\textit{-0.4}) & 22.5 (\textit{-0.8}) \\
VideoChat2 &  38.5 & 38.0 (\textit{-0.5}) & 37.7 (\textit{-0.3})\\
LLaMA-VID & 47.8 & 47.4 (\textit{-0.4}) & 45.9 (\textit{-1.5})\\
GPT-4o &  70.3 & 66.9 (\textit{-3.4}) &66.9 (\textit{-0.0})\\ \hline
\end{tabular}
}
    \end{center}
    \caption{Model Accuracy at Different Video Lengths.}
    \label{duration}
\end{table}

\begin{table}[t]
    \begin{center}
\begin{tabular}{c|c|c}
\hline
Models & Input & Avg. Acc. (\%) \\  \hline
\multirow{2}{*}{MovieChat} &  16 frames & 24.4 \\ 
& 64 frames &  26.0 (\textit{+1.6}) \\ \hline
\multirow{2}{*}{LLaMA-VID} & 16 frames & 46.5 \\ 
 &  64 frames & 46.7 (\textit{+0.2}) \\ \hline
\multirow{2}{*}{TimeChat} & 16 frames & 47.1 \\ 
 & 64 frames &  47.7 (\textit{+0.6}) \\ \hline
\multirow{2}{*}{GPT-4o} &  16 frames & 68.0 \\ 
 &  64 frames & 68.5 (\textit{+0.5})\\ \hline
\end{tabular}
    \end{center}
    \caption{Model accuracy with different numbers of input frames.}
    \label{frames}
\end{table}

\paragraph{The impact of the number of input frames.}
Due to the limitations in context length of some open-source models, the comparison experiments in Tab.~\ref{experiment1} uniformly set the number of input frames to 16. To assess the impact of increasing the number of input frames on model accuracy, we select four models that support longer context inputs. Since TimeChat does not support 128-frame input locally, we ultimately set the number of input frames to 64 and compare the accuracy with that of 16 frames. Detailed results are shown in Tab.~\ref{frames}. 

As the number of input frames increases, the average accuracy of all models improves. This improvement is mainly due to the additional video information provided, which allows the models to capture more details and avoid incorrect answers caused by insufficient information. However, we also observe that with 64 frames, the accuracy only improves slightly. This is partly because 64 frames still miss a significant amount of information in hour-long videos, and partly because, although long-video MLLMs can support longer context inputs, they have not yet fully utilized this information, leaving room for further improvement.

\section{Conclusion}
In this paper, we design a comprehensive and challenging benchmark called ALLVB based on the 9 major existing video understanding tasks, combined with GPT-4o's automated long-video annotation pipeline. In the realm of long-video understanding benchmarks, ALLVB stands out with the largest number of videos, the longest average video duration, and the highest number of Q\&As. Our tests on various MLLMs reveal that existing models still have significant room for improvement in long-video understanding. We hope this benchmark can serve as an objective evaluation metric for future MLLMs and contribute to the advancement of the entire video understanding community.

\paragraph{Limitations}
The videos in ALLVB are all sourced from movies, which already offer a diverse range of content. However, in the future, we hope to incorporate more types of videos, such as sports videos, documentaries, and others. Additionally, the current video Q\&A content is still relatively simple compared to human capabilities. We are considering adding instructional videos, course videos, and other specialized content to assess MLLMs' understanding of professional knowledge. We will continue to maintain and expand ALLVB, striving to contribute further to the realization of true AGI.

\section{Acknowledgments}
This work is supported by the National Natural Science Foundation of China (62172155, 62402171, 62402505, 62472434), and the Science and Technology Innovation Program of Hunan Province (2022RC3061).

\bibliography{aaai25}

\clearpage
\setcounter{page}{1}

\appendix

\section{A Datasheet}
\subsection{A.1 Motivation}
\paragraph{\textit{For what purpose was the dataset created?}} \textit{Was there a specific task in mind? Was there a specific gap that needed to be filled? Please provide a description.}

Currently, the mainstream video question-answering benchmarks are still primarily focused on short videos, most of which are under 5 minutes long. As LLM capabilities continue to improve, there is an urgent need for a more comprehensive and challenging long video question-answering dataset.

\paragraph{\textit{Who created the dataset (e.g., which team, research group) and on behalf of which entity (e.g., company, institution, organization)?}} \textit{}

All the authors of this paper, along with recruited volunteers, collaboratively built this dataset. Due to the consideration of anonymity, the organization's names cannot be disclosed at this time.

\paragraph{\textit{Who funded the creation of the dataset?}} \textit{If there is an associated grant, please provide the name of the grantor and the grant name and number.}

No grant.

\paragraph{\textit{Any other comments?}}\textit{}

None.

\subsection{A.2 Composition}
\paragraph{\textit{What do the instances that comprise the dataset represent (e.g., documents, photos, people, countries)?}}\textit{Are there multiple types of instances (e.g., movies, users, and ratings; people and interactions between them; nodes and edges)? Please provide a description.}

Each instance in the dataset includes a feature-length movie, the corresponding subtitles, and the corresponding Q\&As. Each Q\&A consists of five options.

\paragraph{\textit{How many instances are there in total (of each type, if appropriate)?}}\textit{}

There are a total of 1,376 movies, with a combined duration of 2,628 hours, and a total of 252,420 Q\&As.

\paragraph{\textit{Does the dataset contain all possible instances or is it a sample (not necessarily random)
of instances from a larger set?}}\textit{If the dataset is a sample, then what is the larger set? Is the sample representative of the larger set (e.g., geographic coverage)? If so, please describe how this representativeness was validated/verified. If it is not representative of the larger set, please describe why not (e.g., to cover a more diverse range of instances, because instances were withheld or unavailable).}

We carefully selected 1,376 movies from all publicly available films based on their quality and variety. The dataset included 16 main genres of movies, which essentially represented all common movie categories.

\paragraph{\textit{What data does each instance consist of?}}\textit{``Raw'' data (e.g., unprocessed text or images) or features? In either case, please provide a description.}

Each instance includes one movie, with an average length of 114 minutes, and an average of 183 Q\&As. Each Q\&A contains five options, with one being the correct answer.

\paragraph{\textit{Is there a label or target associated with each instance?}}\textit{If so, please provide a description.}

Yes, each Q\&A for the movie includes one correct option: A/B/C/D/E.

\paragraph{\textit{Is any information missing from individual instances?}}\textit{If so, please provide a description, explaining why this information is missing (e.g., because it was unavailable). This does not include intentionally removed information, but might include, e.g., redacted text.}

There is no missing information; all instances are complete and self-consistent.

\paragraph{\textit{Are relationships between individual instances made explicit (e.g., users’ movie ratings,
social network links)?}}\textit{If so, please describe how these relationships are made explicit.}

There is no obvious relationship between the instances.

\paragraph{\textit{Are there recommended data splits (e.g., training, development/validation, testing)?}}\textit{If so,
please provide a description of these splits, explaining the rationale behind them.}

We divided the dataset into a training set and a test set in a 9:1 ratio. The test set can be used directly to test the long video understanding capabilities of MLLMs, while the training set can be used for further pre-training or fine-tuning of MLLMs.

\paragraph{\textit{Are there any errors, sources of noise, or redundancies in the dataset?}}\textit{If so, please provide a description.}

Since the video is first divided into different subplots, the same type of questions may be asked for different subplots based on their content, such as objects, actions, and so on.

\paragraph{\textit{Is the dataset self-contained, or does it link to or otherwise rely on external resources (e.g.,
websites, tweets, other datasets)?}}\textit{ If it links to or relies on external resources, a) are there
guarantees that they will exist, and remain constant, over time; b) are there official archival
versions of the complete dataset (i.e., including the external resources as they existed at the
time the dataset was created); c) are there any restrictions (e.g., licenses, fees) associated
with any of the external resources that might apply to a dataset consumer? Please provide
descriptions of all external resources and any restrictions associated with them, as well as
links or other access points, as appropriate.}

Yes, the dataset is self-contained, and all the movies included in the dataset are publicly accessible. We will upload all the movies, subtitles, and Q\&A files as soon as possible for use under the CC-BY-NC-SA-4.0 license.

\paragraph{\textit{Does the dataset contain data that might be considered confidential (e.g., data that is
protected by legal privilege or by doctor–patient confidentiality, data that includes the content of individuals’ non-public communications)?}}\textit{If so, please provide a description.}

No.

\paragraph{\textit{Does the dataset contain data that, if viewed directly, might be offensive, insulting, threatening, or might otherwise cause anxiety?}}\textit{ If so,
please describe why.}

No.

\paragraph{\textit{Does the dataset identify any subpopulations (e.g., by age, gender)?}}\textit{If so, please describe how these subpopulations are identified and provide a description of their respective distributions within the dataset.}

No.

\paragraph{\textit{Is it possible to identify individuals (i.e., one or more natural persons), either directly or indirectly (i.e., in combination with other data) from the dataset?}}\textit{If so, please describe how.}

No.

\paragraph{\textit{Does the dataset contain data that might be considered sensitive in any way (e.g., data that reveals race or ethnic origins, sexual orientations, religious beliefs, political opinions or union memberships, or
locations; financial or health data; biometric or genetic data; forms of government identification, such as social security numbers; criminal history)?}}\textit{If so, please provide a description.}

No.

\paragraph{\textit{Any other comments?}}\textit{}

No.

\subsection{A.3 Collection Process}
\paragraph{\textit{How was the data associated with each instance acquired?}}\textit{Was the
data directly observable (e.g., raw text, movie ratings), reported by subjects (e.g., survey responses), or indirectly inferred/derived from other data (e.g., part-of-speech tags, model-based guesses for age or language)? If the data was reported by subjects or indirectly inferred/derived from other data, was the data validated/verified? If so, please describe how.}

The movies and subtitles are downloaded from open-source websites using scripts. The Q\&As are automatically generated using the GPT-4 API and are ultimately validated manually to obtain the final dataset.

\paragraph{\textit{What mechanisms or procedures were used to collect the data (e.g., hardware apparatuses or sensors, manual human curation, software programs, software APIs)?}}\textit{How were these mechanisms or procedures
validated?}

We use web crawlers to collect movies and subtitles from open-source video and subtitle websites. Then, we utilize the GPT-4 API to generate the Q\&As automatically.

\paragraph{\textit{If the dataset is a sample from a larger set, what was the sampling strategy (e.g., deterministic, probabilistic with specific sampling probabilities)?}}\textit{}

Please refer to the third question in section A.2.

\paragraph{\textit{Who was involved in the data collection process (e.g., students, crowdworkers, contractors) and how were they compensated (e.g., how much were crowdworkers paid)?}}\textit{}

Since we use LLM for automatic annotation, we do not hire crowd workers. However, all collaborators on the dataset, along with some recruited volunteers, participate in the dataset verification process.

\paragraph{\textit{Over what timeframe was the data collected?}}\textit{Does this timeframe
match the creation timeframe of the data associated with the instances (e.g., recent crawl of old news articles)? If not, please describe the timeframe in which the data associated with the instances was created.}

The dataset is collected from June to August 2024.

\paragraph{\textit{Were any ethical review processes conducted (e.g., by an institutional review board)?}}\textit{If so, please provide a description of these review processes, including the outcomes, as well as a link or other access point to any supporting documentation.}

No.

\paragraph{\textit{Did you collect the data from the individuals in question directly, or obtain it via third parties or other sources (e.g., websites)?}}\textit{}

The data was directly downloaded from open-source websites.

\paragraph{\textit{Were the individuals in question notified about the data collection?}}\textit{If so, please describe (or show with screenshots or other information) how
notice was provided, and provide a link or other access point to, or otherwise reproduce, the exact language of the notification itself.}

The downloaded movies and subtitles can be freely used for educational purposes.

\paragraph{\textit{Did the individuals in question consent to the collection and use of their data?}}\textit{ If so, please describe (or show with screenshots or other
information) how consent was requested and provided, and provide a link or other access point to, or otherwise reproduce, the exact language to which the individuals consented.}

No (See previous question).

\paragraph{\textit{If consent was obtained, were the consenting individuals provided with a mechanism to revoke their consent in the future or for certain uses?}}\textit{If so, please provide a description, as well as a link or other access point to the mechanism (if appropriate).}

N/A.

\paragraph{\textit{Has an analysis of the potential impact of the dataset and its use on data subjects (e.g., a data protection impact analysis) been conducted?}}\textit{ If so, please provide a description of this analysis, including the
outcomes, as well as a link or other access point to any supporting documentation.}

N/A.

\paragraph{\textit{Any other comments?}}\textit{}

No.

\subsection{A.4 Preprocessing/cleaning/labeling}
\paragraph{\textit{Was any preprocessing/cleaning/labeling of the data done (e.g., discretization or bucketing, tokenization, part-of-speech tagging, SIFT
feature extraction, removal of instances, processing of missing values)?}}\textit{If so, please provide a description. If not, you may skip the remaining questions in this section.}

We screened the movies and their corresponding scripts to ensure high-quality content and to exclude any sensitive or harmful material.

\paragraph{\textit{Was the “raw” data saved in addition to the preprocessed/cleaned/labeled data (e.g., to support unanticipated future uses)?}}\textit{ If so, please provide a link or other access point to the “raw” data.}

No, we did not retain the raw data before preprocessing.

\paragraph{\textit{Is the software that was used to preprocess/clean/label the data available?}}\textit{If so, please provide a link or other access point.}

We wrote Python scripts for data cleaning based on specific rules. These rules are explained in the construction pipeline section of Part 3 in the paper. We have not publicly released these scripts yet, but we can provide them if needed.

\paragraph{\textit{Any other comments?}}\textit{}

No.

\subsection{A.5 Uses}
\paragraph{\textit{Has the dataset been used for any tasks already?}}\textit{If so, please provide a description.}

As of August 2024, only the original paper has used this dataset.

\paragraph{\textit{Is there a repository that links to any or all papers or systems that
use the dataset?}}\textit{If so, please provide a link or other access point.}

Currently, the original paper and code have not been open-sourced, but we will provide the open-source links as soon as possible.

\paragraph{\textit{What (other) tasks could the dataset be used for?}}\textit{}

This dataset is primarily used for multi-modal video QA tasks. When constructing the dataset, we based our question templates on existing video understanding tasks, including Video Classification, Video Scene Recognition, Video Action Recognition, Needle-in-a-Haystack task for video, and so on. Therefore, this dataset can comprehensively assess the understanding capabilities of MLLMs on long videos.

\paragraph{\textit{Is there anything about the composition of the dataset or the way it was collected and preprocessed/cleaned/labeled that might impact future uses?}}\textit{For example, is there anything that a dataset consumer might need to know to avoid uses that could result in unfair treatment of individuals or groups (e.g., stereotyping, quality of service issues) or other
risks or harms (e.g., legal risks, financial harms)? If so, please provide a description. Is there anything a dataset consumer could do to mitigate these risks or harms?}

No.

\paragraph{\textit{Are there tasks for which the dataset should not be used?}}\textit{If so, please provide a description.}

No.

\paragraph{\textit{Any other comments?}}\textit{}

No.

\subsection{A.6 Distribution}
\paragraph{\textit{Will the dataset be distributed to third parties outside of the entity (e.g., company, institution, organization) on behalf of which the dataset was created?}}\textit{If so, please provide a description.}

Yes, we plan to make the dataset publicly available.

\paragraph{\textit{How will the dataset will be distributed (e.g., tarball on website, API, GitHub)?}}\textit{Does the dataset have a digital object identifier (DOI)?}

We plan to upload the dataset in JSON format to HuggingFace. This dataset does not have a DOI.

\paragraph{\textit{When will the dataset be distributed?}}\textit{}

The dataset will be released in 2024.

\paragraph{\textit{Will the dataset be distributed under a copyright or other intellectual property (IP) license, and/or under applicable terms of use (ToU)?}}\textit{If
so, please describe this license and/or ToU, and provide a link or other access point to, or otherwise reproduce, any relevant licensing terms or ToU, as well as any fees associated with these restrictions.}

Our dataset is under the CC-BY-NC-SA-4.0 license.

\paragraph{\textit{Have any third parties imposed IP-based or other restrictions on the data associated with the instances?}}\textit{If so, please describe these restrictions, and provide a link or other access point to, or otherwise reproduce, any relevant licensing terms, as well as any fees associated with these restrictions.}

No.

\paragraph{\textit{Do any export controls or other regulatory restrictions apply to the dataset or to individual instances?}}\textit{ If so, please describe these restrictions, and provide a link or other access point to, or otherwise reproduce, any supporting documentation.}

No.

\paragraph{\textit{Any other comments?}}\textit{}

No.

\subsection{Maintenance}
\paragraph{\textit{Who will be supporting/hosting/maintaining the dataset?}}\textit{}

We will upload the dataset to Huggingface, and the authors of the paper will be responsible for its subsequent maintenance.

\paragraph{\textit{How can the owner/curator/manager of the dataset be contacted (e.g., email address)?}}\textit{}

The authors' email addresses will be provided on the first page of the paper.

\paragraph{\textit{Is there an erratum?}}\textit{If so, please provide a link or other access point.}

No.

\paragraph{\textit{Will the dataset be updated (e.g., to correct labeling errors, add new instances, delete instances)?}}\textit{If so, please describe how often, by
whom, and how updates will be communicated to dataset consumers (e.g., mailing list, GitHub)?}

Yes, the authors of this paper will continue to add new instances or correct any potential errors in the future. Any significant updates will be notified on HuggingFace.

\paragraph{\textit{If the dataset relates to people, are there applicable limits on the retention of the data associated with the instances (e.g., were the individuals in question told that their data would be retained for a fixed
period of time and then deleted)?}}\textit{If so, please describe these limits and explain how they will be enforced.}

No.

\paragraph{\textit{Will older versions of the dataset continue to be supported/hosted/maintained?}}\textit{If so, please describe how. If not, please describe how its obsolescence will be communicated to dataset consumers.}

N/A. There are no older versions at the current moment. All updates regarding the current version will be communicated via HuggingFace.

\paragraph{\textit{If others want to extend/augment/build on/contribute to the dataset, is there a mechanism for them to do so?}}\textit{If so, please provide a description. Will these contributions be validated/verified? If so, please describe how. If not, why not? Is there a process for communicating/distributing these contributions to dataset consumers? If so, please provide a description.}

We do not currently plan to grant third parties permission to modify the dataset. This approach ensures the consistency and fairness of the dataset.

\paragraph{\textit{Any other comments?}}\textit{}

No.

\section{B Further discussion on GPT-4o}
Given that the benchmark was annotated using GPT-4o, there may be concerns regarding the fairness of the experiments. Here are a few points that need further clarification:
\begin{itemize}
    \item Although the Q\&As were annotated using GPT-4o, the annotations were based on movie scripts, while the testing was conducted using the movies themselves. Therefore, during testing, all models have no access to the original script information and must rely on their video understanding capabilities to answer the questions.
    \item After GPT-4o completed the annotations, multiple rounds of verification and extensive revisions were carried out during the quality control phase for Q\&As that did not meet the required standards. As a result, the Q\&As contents now differ significantly from the initial annotations.
    \item All tests were conducted in a 0-shot manner, ensuring there was no data leakage or testing on the training dataset. All models encountered the questions for the first time during testing, guaranteeing fairness.
\end{itemize}

\newpage
\onecolumn
\section{C More Q\&A Examples}
Below are the examples of 91 sub-tasks under the 9 main tasks. The correct options are in bold green.
\subsection{C.1 Sub-tasks for Video Classification (VC)}
\begin{figure*}[ht]
    \begin{center}
    \centerline{\includegraphics[width=1\linewidth]{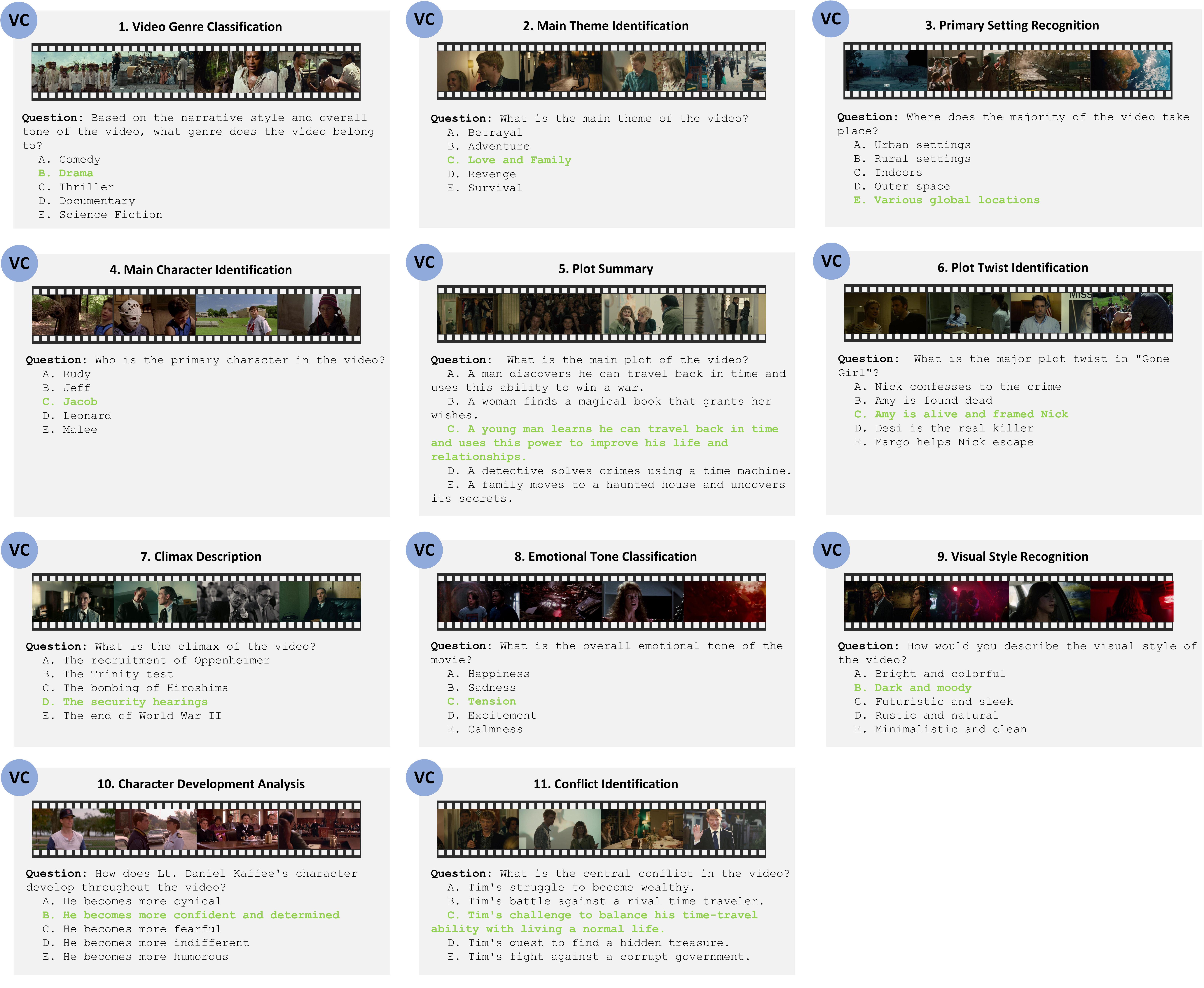}}
     \caption{Examples of the 11 sub-tasks for the Video Classification (VC) task.}
     \label{vc}
     \end{center}
\end{figure*}

\newpage
\subsection{C.2 Sub-tasks for Scene Recognition (SR)}
\begin{figure*}[ht]
    \begin{center}
    \centerline{\includegraphics[width=1\linewidth]{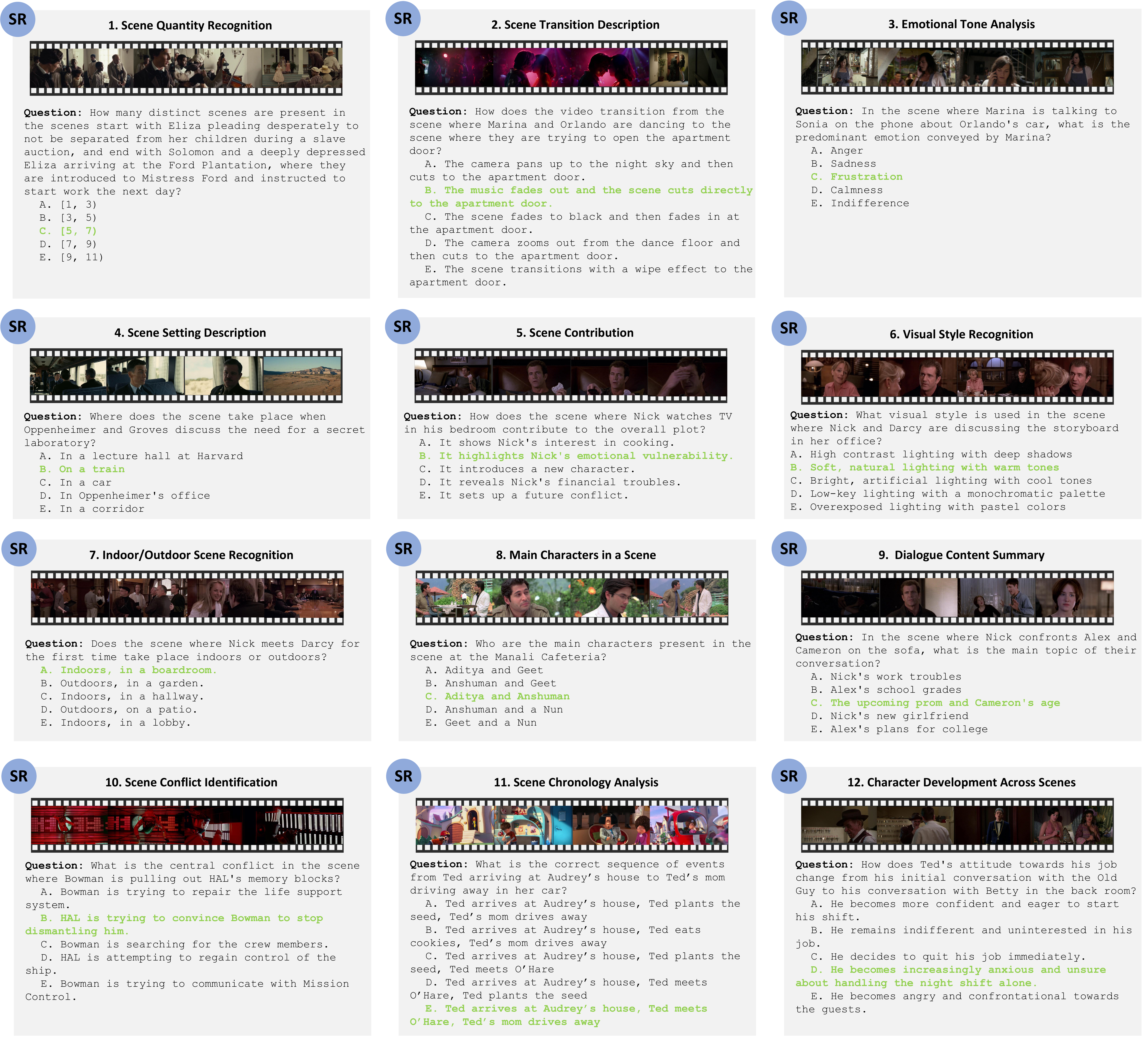}}
     \caption{Examples of the 12 sub-tasks for the Scene Recognition (SR) task.}
     \label{sr}
     \end{center}
\end{figure*}

\newpage
\subsection{C.3 Sub-tasks for Object Detection and Tracking (ODT)}
\begin{figure*}[ht]
    \begin{center}
    \centerline{\includegraphics[width=0.9\linewidth]{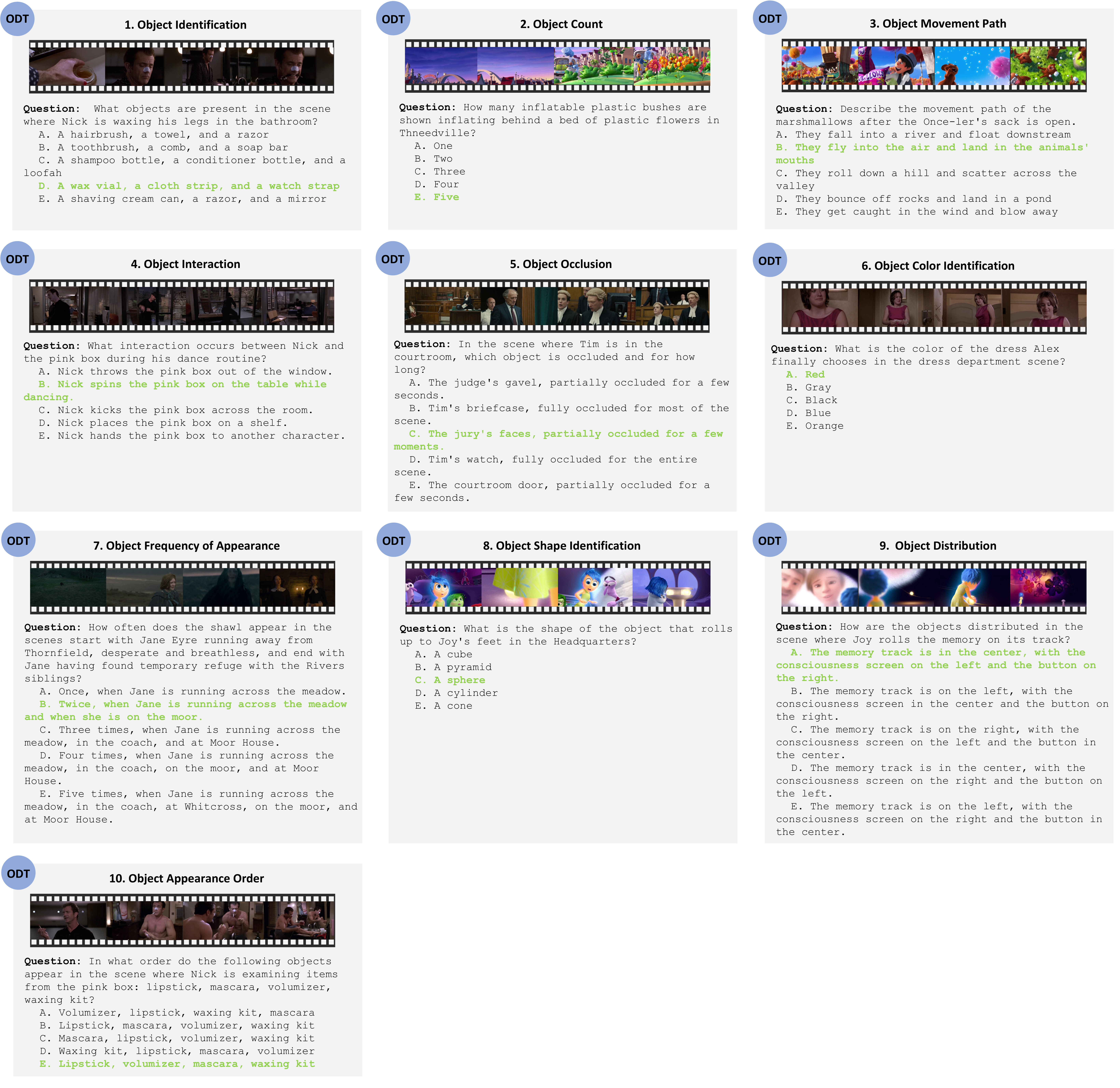}}
     \caption{Examples of the 10 sub-tasks for the Object Detection and Tracking (ODT) task.}
     \label{odt}
     \end{center}
\end{figure*}

\newpage
\subsection{C.4 Sub-tasks for Action Recognition (AR)}
\begin{figure*}[ht]
    \vskip -0.1in
    \begin{center}
    \centerline{\includegraphics[width=1\linewidth]{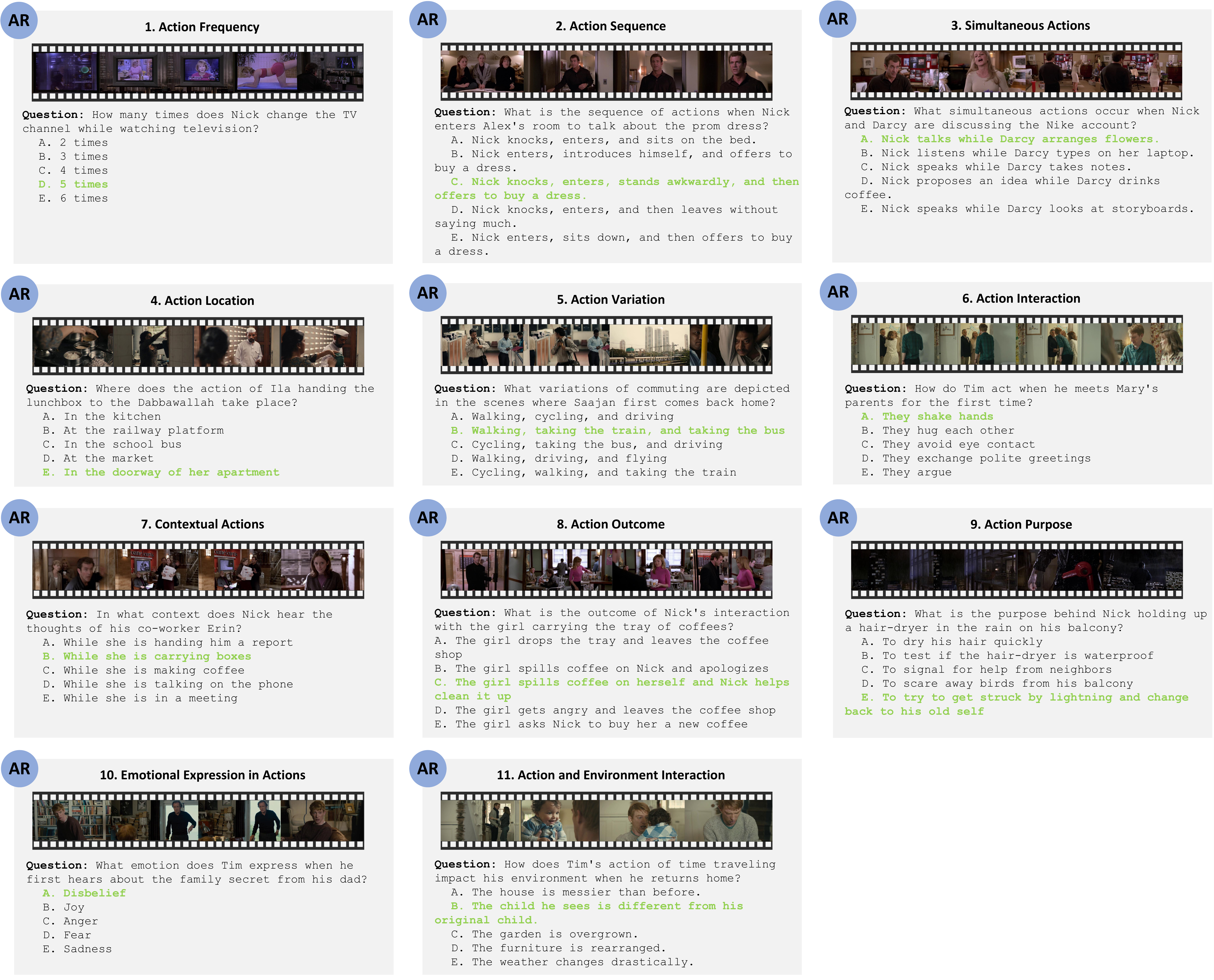}}
     \caption{Examples of the 11 sub-tasks for the Action Recognition (AR) task.}
     \label{ar}
     \end{center}
\end{figure*}

\newpage
\subsection{C.5 Sub-tasks for Temporal Action Localization (TAL)}
\begin{figure*}[ht]
    \vskip -0.1in
    \begin{center}
    \centerline{\includegraphics[width=1\linewidth]{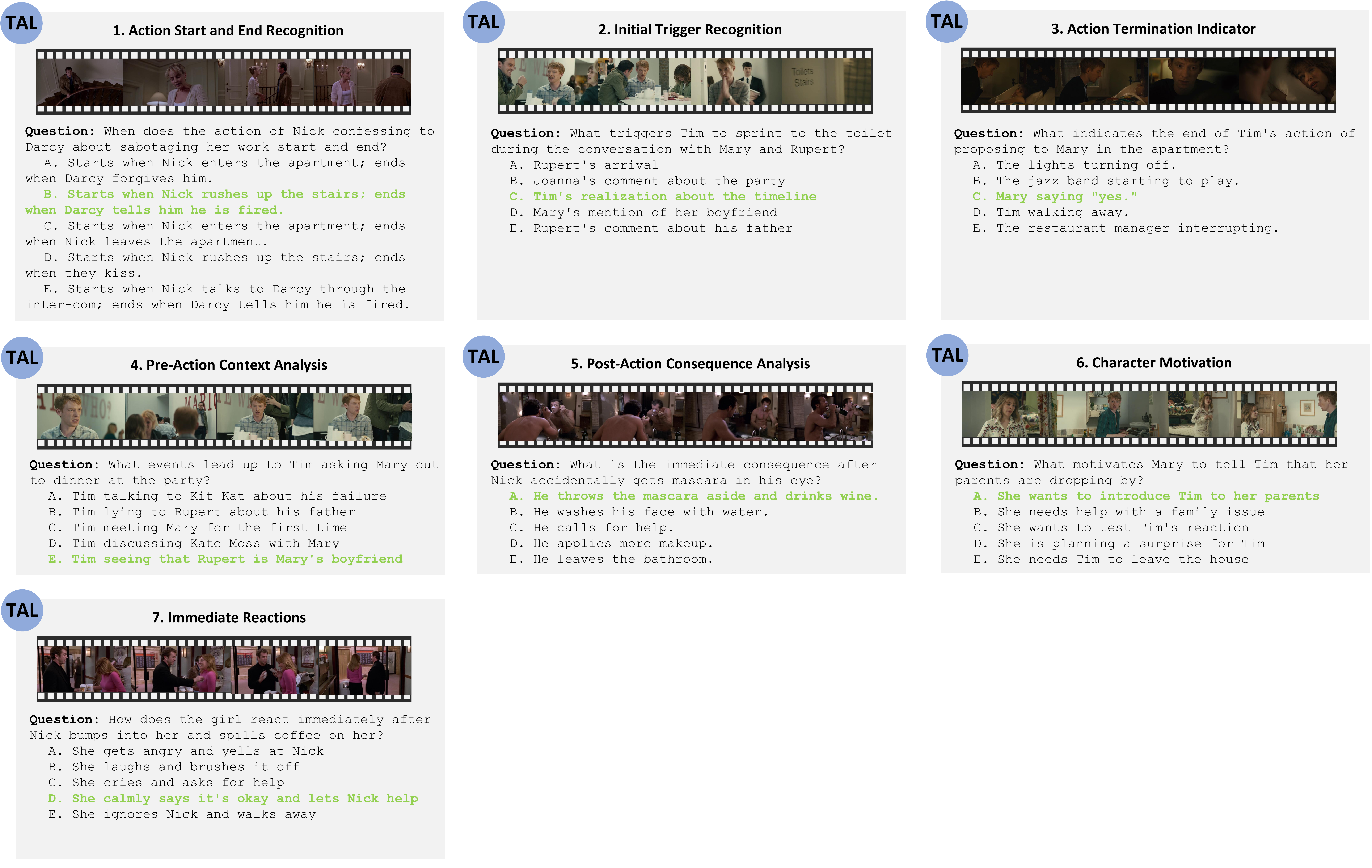}}
     \caption{Examples of the 7 sub-tasks for the Temporal Action Localization (TAL) task.}
     \label{tal}
     \end{center}
\end{figure*}

\newpage
\subsection{C.6 Sub-tasks for Event Detection (ED)}
\begin{figure*}[ht]
    \vskip -0.1in
    \begin{center}
    \centerline{\includegraphics[width=1\linewidth]{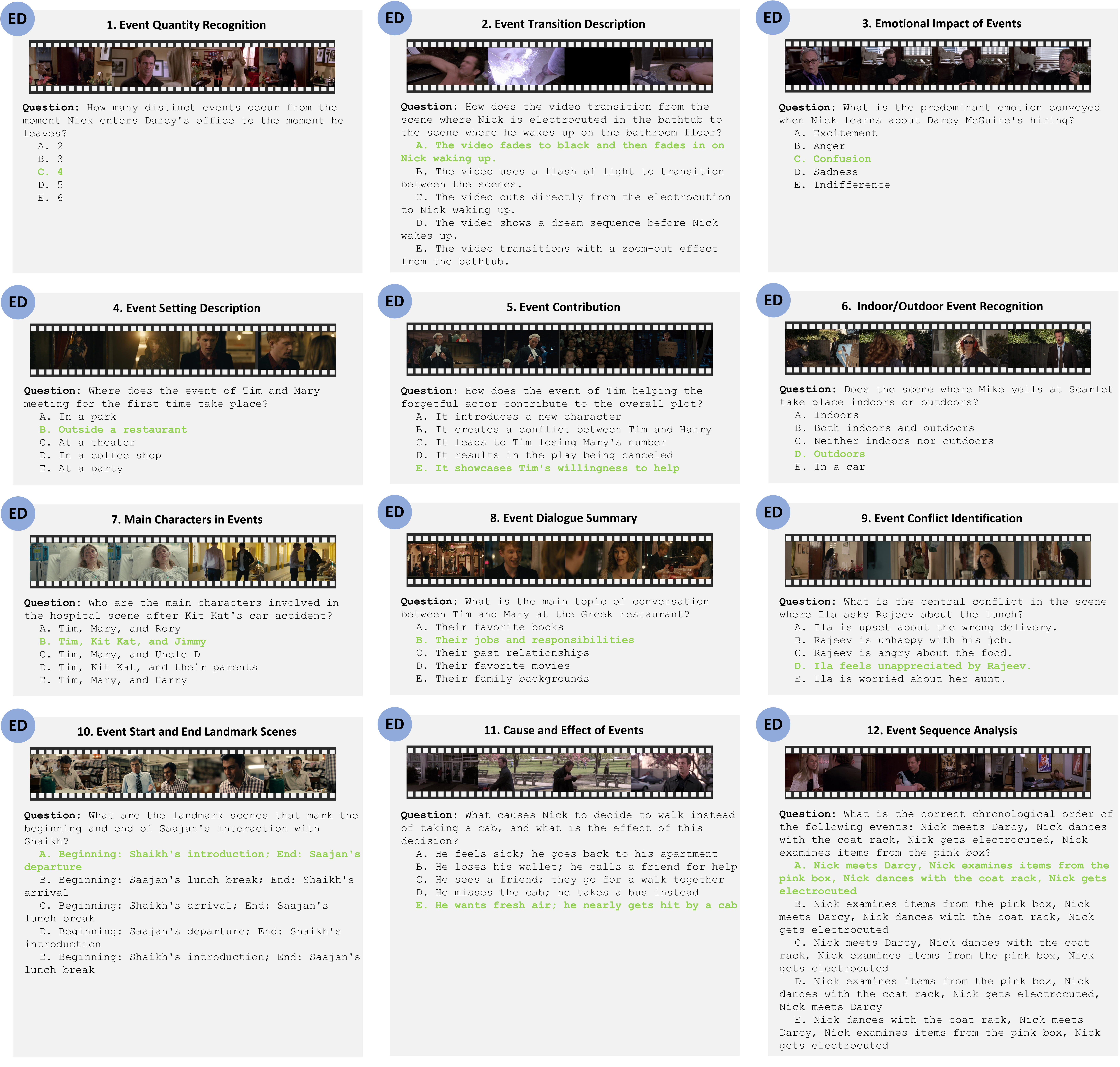}}
     \caption{Examples of the 12 sub-tasks for the Event Detection (ED) task.}
     \label{ed}
     \end{center}
\end{figure*}

\newpage
\subsection{C.7 Sub-tasks for Video Captioning (VCap)}
\begin{figure*}[ht]
    \vskip -0.1in
    \begin{center}
    \centerline{\includegraphics[width=1\linewidth]{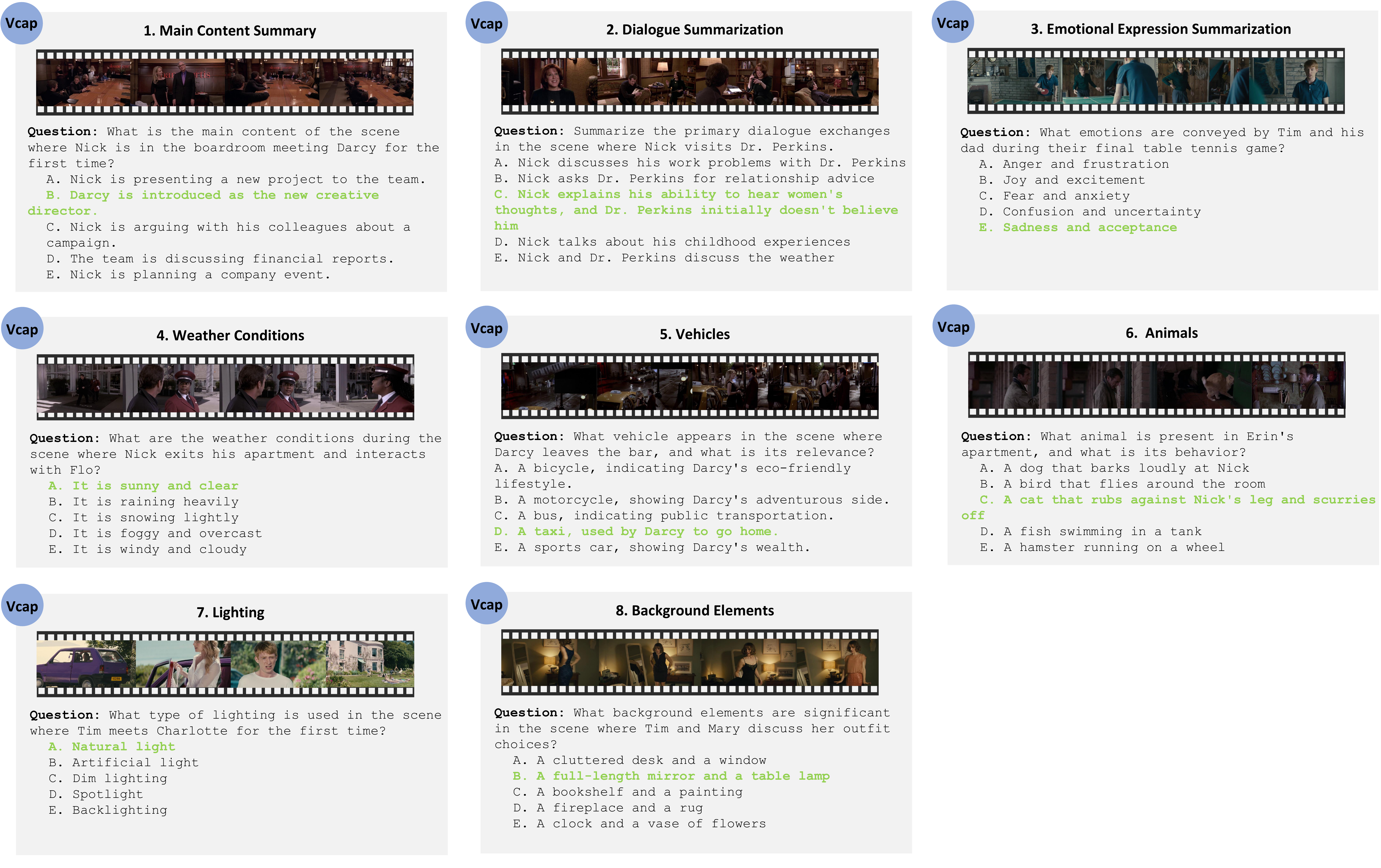}}
     \caption{Examples of the 8 sub-tasks for the Video Captioning (VCap) task.}
     \label{vcap}
     \end{center}
\end{figure*}

\newpage
\subsection{C.8 Sub-tasks for Video Emotion Recognition (VER)}
\begin{figure*}[ht]
    \vskip -0.1in
    \begin{center}
    \centerline{\includegraphics[width=1\linewidth]{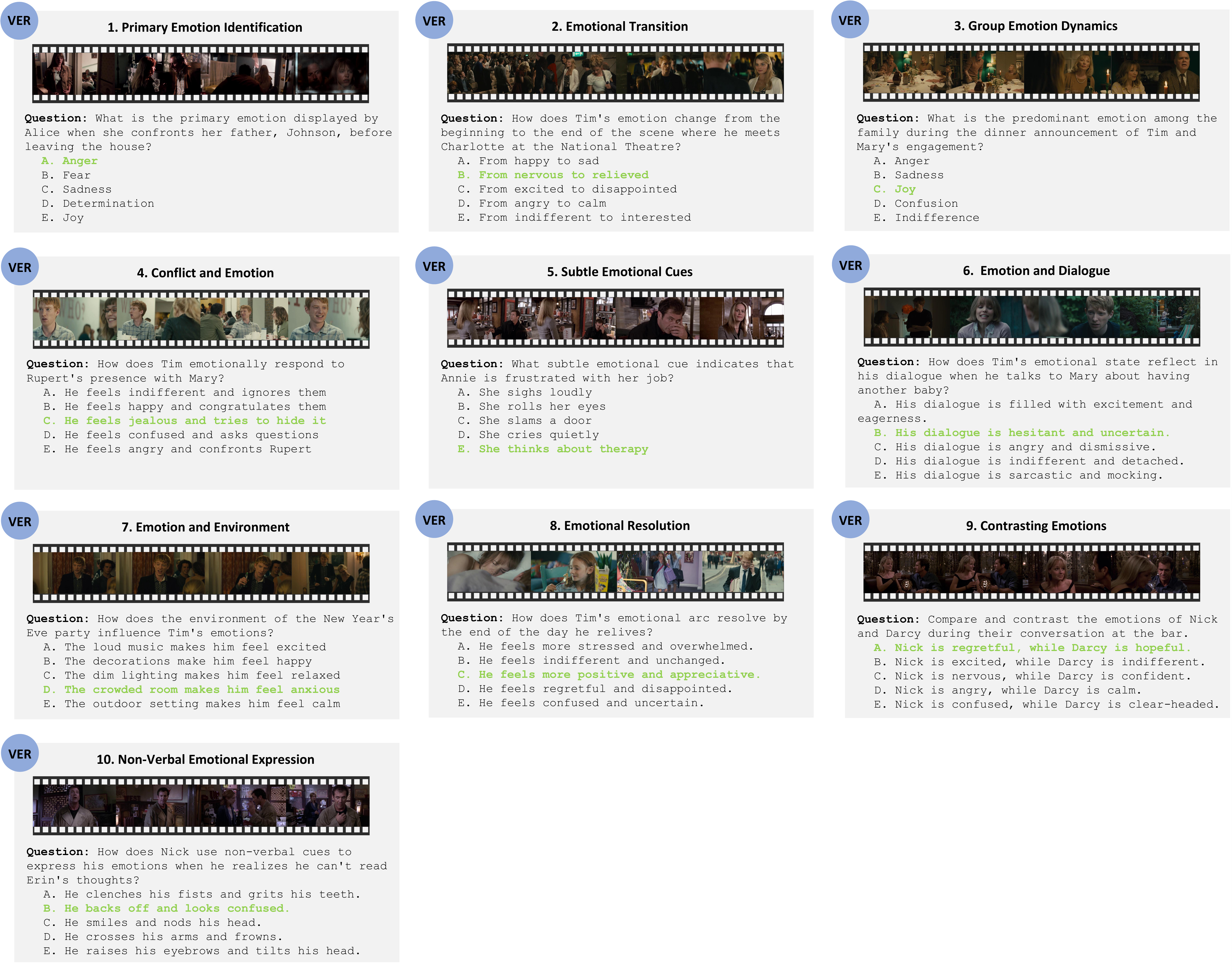}}
     \caption{Examples of the 10 sub-tasks for the Video Emotion Recognition (VER) task.}
     \label{ver}
     \end{center}
\end{figure*}

\newpage
\subsection{C.9 Sub-tasks for Needle-in-a-Haystack (NH)}
\begin{figure*}[ht]
    \vskip -0.1in
    \begin{center}
    \centerline{\includegraphics[width=0.9\linewidth]{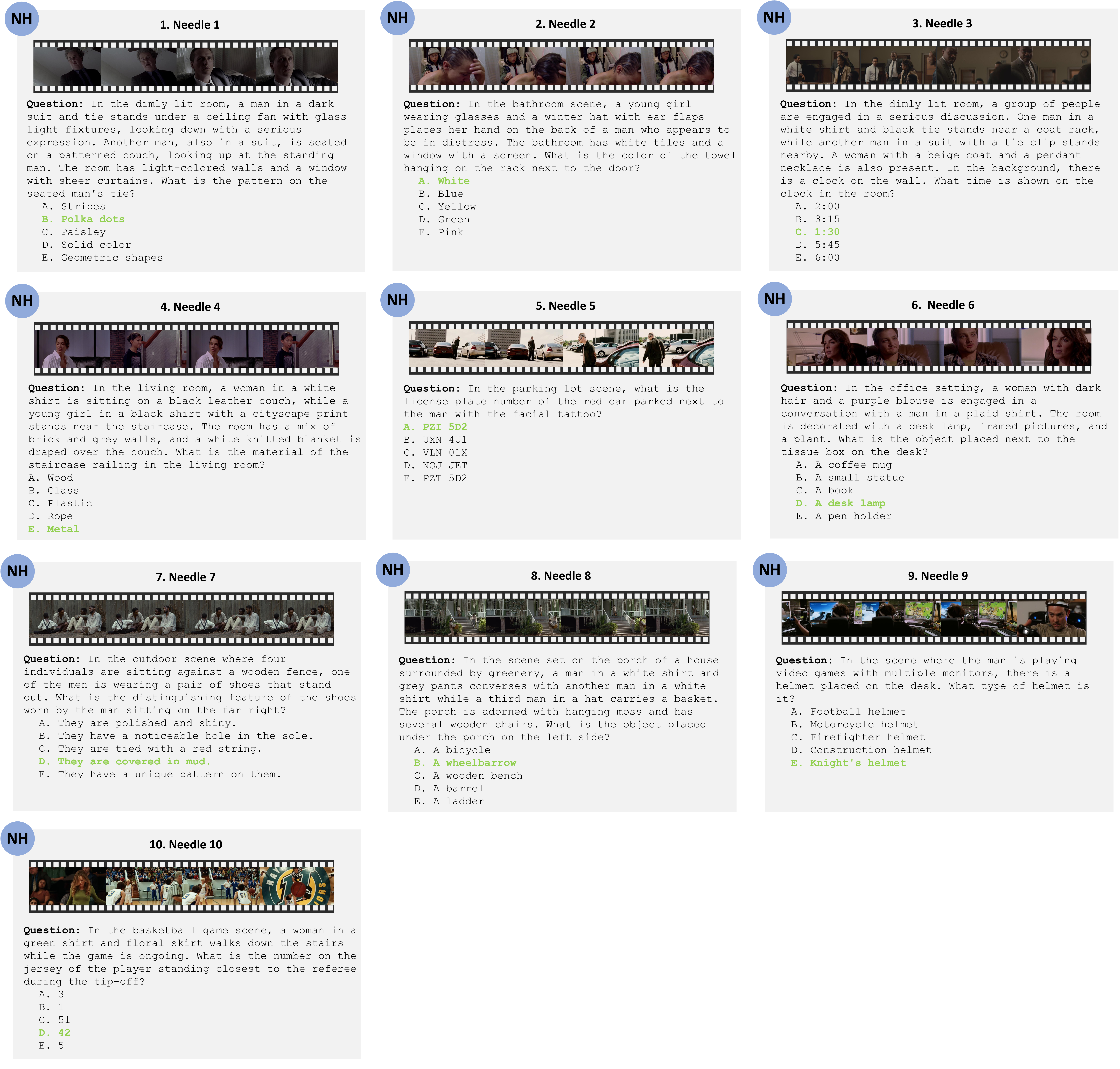}}
     \caption{Examples of the 10 sub-tasks for the Needle-in-a-Haystack (NH) task.}
     \label{nh}
     \end{center}
\end{figure*}

\newpage
\section{D Prompt}
\subsection{D.1 Plot Segmentation Prompt}
The following prompts instruct GPT-4o to first use a two-stage segmentation method to divide the movie script into different sub-plots and then design questions based on the entire script content and 11 video classification sub-tasks.

\begin{quote}
\begin{scriptsize}

Below is the script of a certain movie or TV series, with each page formatted as follows:

--------------------

P\#number:

[Current page script content]

In this format, the ``\#number'' in ``P\#number'' indicates the page number of the script, while numbers in other positions do not represent page numbers. Please strictly follow this format to determine the total number of pages. Below is the script content:

\{script\_text\}

The script includes descriptions of video scenes, character names, and dialogues between them. To better help others understand the content of this movie or TV series, your task is:

1. Identify the [Script Title], the corresponding [Genre] on IMDB, the start and end [Page Number] of the main content of the script, the [Total Number of Pages] of the main content, and the main [Plot Content] of the script.

2. List in detail all the 1-N plots contained in the script. Clearly indicate the [Plot Title], the start and end [Page Number] of each plot, the [Percentage] of the start and end [Page Number] subtracted by the main content start [Page Number] divided by the [Total Number of Pages] of the main content (written in decimal form, keeping 4 significant digits), and the [Plot Content]. Please strictly follow the example provided below and ensure the [Page Number] matches the ``\#number'' in ``P\#number''.  Reiterate that only the ``\#number'' in ``P\#number'' indicates the page number, and all other numbers do not represent page numbers.

\#\#\# Example

Script Title: [Script Title] 

IMDB Genre: [Genre]  

Main Content Start Page: [Page Number]  

Main Content End Page: [Page Number]  

Total Number of Pages for Main Content: [Total Number of Pages] 

Main Plot Content of the Script: [Plot Content]

Here are the main plot summaries from the document:  

Plot 1 Title: [Plot Title] 

Start Page: [Page Number] - [Percentage]  

End Page: [Page Number] - [Percentage]

Plot Content: [Plot Content]

Plot 2 Title: [Plot Title] 

Start Page: [Page Number] - [Percentage]  

End Page: [Page Number] - [Percentage]

Plot Content: [Plot Content]

[...]

Due to the length of the script, to better understand its content, you need to further divide each of the above plots into 1-M subplots. The start [Page Number] of the first subplot and the end [Page Number] of the last subplot should match the start and end [Page Number] of the main plot. Clearly indicate the [Subplot Title], the start and end [Page Number] of each subplot, the [Percentage] of the start and end [Page Number] subtracted by the main content start [Page Number] divided by the [Total Number of Pages] of the main content (written in decimal form, keeping 4 significant digits), and the [Subplot Content]. 

Please strictly follow the example provided below and ensure the [Page Number] matches the ``\#number'' in ``P\#number''. Reiterate that only the ``\#number'' in ``P\#number'' indicates the page number, and all other numbers do not represent page numbers.

\#\#\# Example

\#\#\#\# Subplots of Plot 1

Subplot 1 Title: [Subplot Title] 

Start Page: [Page Number] - [Percentage]  

End Page: [Page Number] - [Percentage]  

Subplot Content: [Subplot Content]

Subplot 2 Title: [Subplot Title]

Start Page: [Page Number] - [Percentage]  

End Page: [Page Number] - [Percentage]

Subplot Content: [Subplot Content]

[...]

Treat reading the script as if you are watching the video it describes. To better analyze the script, based on the content of the script, please design 11 [Questions] according to [Question Types] and their corresponding [Question Requirements] and [Question Description] provided below. Each question should include 5 options (A, B, C, D, E), with only one correct option. The requirements for correct and incorrect options, as well as the final answer, are as follows:

1. All incorrect options and the correct option should be of similar length, neither too long nor too short.

2. Incorrect options should be clearly different from the correct option but closely related to the video content.

3. The correct option should not be answerable based solely on the question content; it should require information from the video content.

4. The correct option should be randomly distributed among the incorrect options.

5. The answer should include the letter of the [Correct option] (A, B, C, D, E) and the [reasoning process] for how this answer was determined.

The [Questions] templates are as follows:

\#\#\#\# Question Type 1: [Question Type]

\*\*Question:\*\* Briefly describe the selected scenes and ask the question.

- A. [Option content]

- B. [Option content]

- C. [Option content]

- D. [Option content]

- E. [Option content]

\*\*Answer\*\*: [Correct option] + [reasoning process]

[...]

[Question Types] as follows:

\#\#\#\# Question Type 1: Video Genre Classification (VC)

**Question Requirements:** Identify the genre of the video.

**Question Description:** Determine the genre (e.g., comedy, drama, thriller, documentary) of the video based on its visual and auditory cues, narrative style, and overall tone.

\#\#\#\# Question Type 2: Main Theme Identification (VC)

**Question Requirements:** Determine the main theme of the video.  

**Question Description:** Identify the central theme or message conveyed in the video, such as friendship, betrayal, adventure, or love, based on the storyline and character interactions.

\#\#\#\# Question Type 3: Primary Setting Recognition (VC)

**Question Requirements:** Identify the primary setting of the video.  

**Question Description:** Describe the main location where the video takes place, including details such as whether it is an urban or rural setting, indoors or outdoors, and any notable landmarks or environments.

\#\#\#\# Question Type 4: Main Character Identification (VC)

**Question Requirements:** Identify the main character(s) in the video.  

**Question Description:** Determine who the primary character(s) are in the video, focusing on their role, appearance, and actions.

\#\#\#\# Question Type 5: Plot Summary (VC)

**Question Requirements:** Provide a brief summary of the video’s plot.

**Question Description:** Summarize the key events and actions that occur in the video, highlighting the main storyline and important developments.

\#\#\#\# Question Type 6: Plot Twist Identification (VC)

**Question Requirements:** Identify any plot twists in the video. 

**Question Description:** Describe the unexpected events or revelations that change the direction of the story and explain their impact on the characters and narrative.

\#\#\#\# Question Type 7: Climax Description (VC)

**Question Requirements:** Identify and describe the climax of the video.  

**Question Description:** Describe the most intense or critical point of the story, where the main conflict reaches its peak, and explain its significance.

\#\#\#\# Question Type 8: Emotional Tone Classification (VC)

**Question Requirements:** Determine the overall emotional tone of the video.  

**Question Description:** Identify the predominant emotion conveyed in the video, such as happiness, sadness, tension, or excitement, based on character expressions, music, and dialogue.

\#\#\#\# Question Type 9: Visual Style Recognition (VC)

**Question Requirements:** Identify the visual style used in the video. 

**Question Description:** Describe the visual elements, such as cinematography, color palette, lighting, and shot composition, that characterize the visual style of the video.

\#\#\#\# Question Type 10: Character Development Analysis (VC)

**Question Requirements:** Analyze the development of a specific character throughout the video. 

**Question Description:** Describe how a particular character changes or evolves from the beginning to the end of the video, providing examples of key moments that illustrate this development.

\#\#\#\# Question Type 11: Conflict Identification (VC)

**Question Requirements:** Identify the central conflict in the video. 

**Question Description:** Describe the main source of conflict, whether internal or external, and explain how it drives the plot and affects the characters.
\end{scriptsize}
\end{quote}

\subsection{D.2 Q\&A Construction Prompt (For sub-plots)}
The following prompts instruct GPT-4o to design questions for the movie's sub-plots based on 14 randomly selected sub-tasks.

\begin{quote}
\begin{scriptsize}
Below is a movie or TV series script:

\{script\_text\}

The script includes descriptions of video scenes, character names, dialogues, and other details. To better help others understand the content of this script, your task is to treat reading the script as if you are watching the video it describes. Based on the video content, design 14 questions according to provided 14 [Question Templates] and their corresponding [Question Requirements] and [Question Description], strictly following the [Question Format], and provide the answers.

Please note, for each question, you must randomly select several scenes (not all scenes) from the video, and then ask question based on the selected scenes. The scenes for each question should not be identical and should be evenly distributed throughout the video. Then, you must provide a brief overview of the scenes and ask the question based on the selected scenes. Each question must include 5 options (A, B, C, D, E), with only one correct option. The requirements for options and answer, are as follows:

1. All incorrect options and the correct option should be of similar length, neither too long nor too short.

2. Incorrect options should be clearly different from the correct option but closely related to the video content.

3. The correct option should not be answerable based solely on the question content; it should require information from the video content.

4. The correct option should be randomly distributed among the incorrect options.

5. The answer should include the letter of the correct option (A, B, C, D, E) and the reasoning process for how this answer was determined.

[Question Format] as follows:

\#\#\#\# Template 1: Dialogue Content Summary (SR)

**Question:** In the scene where Piscano is talking to his mother and brother-in-law, what is the main topic of their conversation?

- A. Piscano's gambling habits

- B. Piscano's trips to Las Vegas

- C. Piscano's distrust of Nance

- D. Piscano's financial troubles

- E. Piscano's relationship with his mother

**Answer**: C. Piscano's distrust of Nance + The main topic of conversation is Piscano's distrust of Nance, as he discusses his concerns about Nance taking money and his frustration with not being able to watch over the money in the count room.

[...]

Below are 14 [Question Templates]. You must design one question for each template based on the given video script and provide the answers. Never ask questions about the entire provided script or video segment; instead, select specific scenes from it. Additionally, the question must never include words like ``segment'', ``script'', ``video'', ``described'' or ``provided'' that represent the entire script.

[Question Templates] as follows:

\{question\_types\}

You must provide 14 complete questions and answers, and never interrupt until all results are fully outputted.
\end{scriptsize}
\end{quote}

The 14 sub-tasks will be composed by randomly selecting 2 sub-tasks from each of the following main tasks.
\begin{quote}
\begin{scriptsize}
\textbf{Scene Recognition:}

    \#\#\#\# Question Type 1: Scene Quantity Recognition (SR)

    **Question Requirements:** Identify the total number of scenes in the video. The scene quantity should be expressed as a range, such as [10, 15). Each range should be clearly distinguishable from other ranges.
    
    **Question Description:** Determine how many distinct scenes are present in the video and describe the content of each scene.

    \#\#\#\# Question Type 2: Scene Transition Description (SR)

    **Question Requirements:** Describe the transition between two scenes. 
    
    **Question Description:** Explain how the video transitions from one scene to another, including any visual or audio cues that indicate the change.

    \#\#\#\# Question Type 3: Emotional Tone Analysis (SR)

    **Question Requirements:** Determine the emotional tone of a scene.  
    
    **Question Description:** Identify the predominant emotion conveyed in a scene, supported by examples of character expressions, dialogue, and background music.
    
    \#\#\#\# Question Type 4: Setting Description (SR)

    **Question Requirements:** Describe the setting of a scene.
    
    **Question Description:** Explain where the scene takes place, including details about the environment, time of day, and background elements.

    \#\#\#\# Question Type 5: Scene Contribution (SR)

    **Question Requirements:** Explain how a scene contributes to the overall plot. 
    
    **Question Description:** Analyze the events in a scene and discuss how they advance the story or reveal important information about the characters or situation.
    
    \#\#\#\# Question Type 6: Visual Style Recognition (SR)

    **Question Requirements:** Identify the visual style used in a scene.  
    
    **Question Description:** Describe the cinematographic techniques, lighting, and color palette used in a scene, and explain how they contribute to the mood and narrative.
     
    \#\#\#\# Question Type 7: Indoor/Outdoor Scene Recognition (SR)

    **Question Requirements:** Determine whether a scene takes place indoors or outdoors. 
    
    **Question Description:** Identify the location setting of a scene (indoor or outdoor) and provide evidence from the video, such as background elements or lighting conditions, to support your answer.
     
    \#\#\#\# Question Type 8: Main Characters in a Scene (SR)
    
    **Question Requirements:** Identify the main characters present in a scene.  
    
    **Question Description:** List the key characters appearing in a specific scene and describe their actions and interactions.
     
    \#\#\#\# Question Type 9: Dialogue Content Summary (SR)

    **Question Requirements:** Summarize the main dialogue in a scene.  
    
    **Question Description:** Provide a brief overview of the key conversations that take place in a scene and explain their significance to the plot or character development.
     
    \#\#\#\# Question Type 10: Scene Conflict Identification (SR)

    **Question Requirements:** Identify the central conflict in a scene.
    
    **Question Description:** Describe the main source of tension or conflict within a scene and discuss how it affects the characters and advances the plot.
     
    \#\#\#\# Question Type 11: Scene Chronology Analysis (SR)

    **Question Requirements:** Analyze the chronological order of events in the scenes. 
    
    **Question Description:** Explain the sequence of events in the video, focusing on how scenes are ordered and their temporal relationships.
     
    \#\#\#\# Question Type 12: Character Development Across Scenes (SR)

    **Question Requirements:** Examine character development over multiple scenes. 
    
    **Question Description:** Describe how a character changes or evolves throughout several scenes, providing examples of significant actions or decisions that highlight their development.
 
 \textbf{Object Detection and Tracking:}
  
    \#\#\#\# Question Type 13: Object Identification (ODT)

    **Question Requirements:** Identify and classify different types of objects.  
    
    **Question Description:** List and classify different types of objects present in the video, providing a description of each type based on their characteristics.
 
    \#\#\#\# Question Type 14: Object Count (ODT)

    **Question Requirements:** Determine the total number of a specific type of object in the video segment. 
    
    **Question Description:** Count the total number of a specified object (e.g., cars, people, animals) appearing throughout the video segment.
   
    \#\#\#\# Question Type 15: Object Movement Path (ODT)

    **Question Requirements:** Track the movement path of a specific object.  
    
    **Question Description:** Describe the movement path of a specified object, noting key locations or changes in direction within the video segment.
    
    \#\#\#\# Question Type 16: Object Interaction (ODT)

    **Question Requirements:** Identify interactions between multiple objects. 
    
    **Question Description:** Describe interactions between specified objects, detailing how they interact (e.g., colliding, exchanging items) and the outcomes of these interactions.
   
    \#\#\#\# Question Type 17: Object Occlusion (ODT)

    **Question Requirements:** Identify instances of object occlusion.  
    
    **Question Description:** Describe instances where specified objects are partially or fully occluded by other objects, noting how long and to what extent they are hidden.
    
    \#\#\#\# Question Type 18: Object Color Identification (ODT)

    **Question Requirements:** Identify the colors of specific objects.  
    
    **Question Description:** Describe the colors of specified objects and note any changes in color throughout the video segment.
     
    \#\#\#\# Question Type 19: Object Frequency of Appearance (ODT)

    **Question Requirements:** Determine the frequency of appearance of specific objects.  
    
    **Question Description:** Count how often specified objects appear in the video segment and describe the intervals of their appearances.
     
    \#\#\#\# Question Type 20: Object Shape Identification (ODT)

    **Question Requirements:** Identify the shapes of specific objects.  
    
    **Question Description:** Describe the shapes of specified objects and note any variations in shape throughout the video segment.
    
    \#\#\#\# Question Type 21: Object Distribution (ODT)

    **Question Requirements:** Determine the spatial distribution of objects within the frame.  
    
    **Question Description:** Describe the positions and arrangement of specified objects within the video frame, noting patterns or clusters.
   
    \#\#\#\# Question Type 22: Object Appearance Order (ODT)

    **Question Requirements:** Identify the order in which objects appear.
    
    **Question Description:** Describe the sequence in which specified objects appear in the video segment, noting the exact order of their appearances.

    \textbf{Action Recognition:}
    
    \#\#\#\# Question Type 23: Action Frequency (AR)

    **Question Requirements:** Determine how many times a specific action occurs in the video. Provide a clear overview of the starting and ending points of the segment in the video where the action is observed.
    
    **Question Description:** Identify and count the number of times a particular action, such as jumping, occurs within the video segment.
     
    \#\#\#\# Question Type 24: Action Sequence (AR)

    **Question Requirements:** Identify the sequence of actions in the video. 
    
    **Question Description:** Determine the order in which different actions occur, such as walking, sitting, and standing up.
     
    \#\#\#\# Question Type 25: Simultaneous Actions (AR)

    **Question Requirements:** Identify actions occurring simultaneously in the video.  
    
    **Question Description:** Detect and describe actions that happen at the same time, such as talking and gesturing.
     
    \#\#\#\# Question Type 26: Action Location (AR)

    **Question Requirements:** Determine where specific actions take place in the video.
    
    **Question Description:** Identify the locations within the video where particular actions, like dancing or eating, occur.
    
    \#\#\#\# Question Type 27: Action Variation (AR)

    **Question Requirements:** Identify variations of a specific action in the video.  
    
    **Question Description:** Describe different forms or styles of the same action, like different types of dance moves.
    
    \#\#\#\# Question Type 28: Action Interaction (AR)

    **Question Requirements:** Identify interactions between characters based on their actions.  
    
    **Question Description:** Describe how characters interact through their actions, such as handshakes, hugs, or arguments.
     
    \#\#\#\# Question Type 29: Contextual Actions (AR)

    **Question Requirements:** Determine the context in which actions occur in the video. 
    
    **Question Description:** Explain the context or setting for specific actions, like cooking in a kitchen or playing in a park.
     
    \#\#\#\# Question Type 30: Action Outcome (AR)

    **Question Requirements:** Identify the outcomes of specific actions in the video. 
    
    **Question Description:** Determine the results or consequences of particular actions, such as a character falling after tripping or smiling after receiving good news.
    
    \#\#\#\# Question Type 31: Action Purpose (AR)

    **Question Requirements:** Determine the purpose behind a specific action in the video.  
    
    **Question Description:** Identify the reason or intention behind an action, such as a character running to catch a bus or waving to greet someone.
     
    \#\#\#\# Question Type 32: Emotional Expression in Actions (AR)

    **Question Requirements:** Identify the emotions expressed through actions in the video.  
    
    **Question Description:** Describe the emotions conveyed by characters through their actions, like hugging to show affection or clenching fists in anger.
     
    \#\#\#\# Question Type 33: Action and Environment Interaction (AR)

    **Question Requirements:** Analyze how actions interact with the environment in the video.  
    
    **Question Description:** Describe how characters' actions are influenced by or impact their surroundings, such as a character adjusting their behavior due to weather conditions or interacting with objects in their environment.

    \textbf{Temporal Action Localization:}
    
    \#\#\#\# Question Type 34: Action Start and End Recognition (TAL)

    **Question Requirements:** Identify the landmark scenes that mark the start and end of a specific action in the video.  
    
    **Question Description:** Determine the specific scenes or moments that indicate the beginning and ending of an action, such as a character entering a room or a car driving away.
     
    \#\#\#\# Question Type 35: Initial Trigger Recognition (TAL)

    **Question Requirements:** Identify the scene or event that triggers the start of a specific action in the video. 
    
    **Question Description:** Determine what event or scene initiates an action, such as a signal, a dialogue, or a change in environment.
     
    \#\#\#\# Question Type 36: Action Termination Indicator (TAL)

    **Question Requirements:** Identify the scene or event that indicates the end of a specific action in the video.  
    
    **Question Description:** Determine what event or scene signifies the conclusion of an action, such as a character leaving the frame, a door closing, or a change in lighting.
     
    \#\#\#\# Question Type 37: Pre-Action Context Analysis (TAL)

    **Question Requirements:** Describe the context and events leading up to the start of an action. 
    
    **Question Description:** Analyze the scenes and events that precede an action to understand the context and conditions that lead to its initiation.
     
    \#\#\#\# Question Type 38: Post-Action Consequence Analysis (TAL)

    **Question Requirements:** Describe the immediate consequences following the end of an action. 
    
    **Question Description:** Analyze the scenes and events that follow an action to understand its immediate impact and outcomes.
     
    \#\#\#\# Question Type 39: Character Motivation (TAL)

    **Question Requirements:** Identify the character’s motivation for performing a specific action.  
    
    **Question Description:** Explain why a character initiates an action, providing context from their dialogue, interactions, or the situation leading up to the action.
     
    \#\#\#\# Question Type 40: Immediate Reactions (TAL)

    **Question Requirements:** Identify the immediate reactions of other characters to an action.  
    
    **Question Description:** Describe how other characters respond right after an action occurs, explaining the immediate impact on their behavior and the scene.

    \textbf{Event Detection:}
    
    \#\#\#\# Question Type 41: Event Quantity Recognition (ED)

    **Question Requirements:** Identify the total number of events in the video. Provide a clear overview of the starting and ending points of the segment in the video where the events are observed.
    
    **Question Description:** Determine how many distinct events are present in the video and describe the content of each event.
     
    \#\#\#\# Question Type 42: Event Transition Description (ED)

    **Question Requirements:** Describe the transition between two events.  
    
    **Question Description:** Explain how the video transitions from one event to another, including any visual or audio cues that indicate the change.
    
    \#\#\#\# Question Type 43: Emotional Impact of Events (ED)

    **Question Requirements:** Determine the emotional impact of an event. 
    
    **Question Description:** Identify the predominant emotion conveyed by an event, supported by examples of character expressions, dialogue, and background music.
     
   \#\#\#\# Question Type 44: Event Setting Description (ED)

    **Question Requirements:** Describe the setting of an event. 
    
    **Question Description:** Explain where the event takes place, including details about the environment, time of day, and background elements.
    
    \#\#\#\# Question Type 45: Event Contribution (ED)

    **Question Requirements:** Explain how an event contributes to the overall plot. 
    
    **Question Description:** Analyze the events and discuss how they advance the story or reveal important information about the characters or situation.
     
    \#\#\#\# Question Type 46: Indoor/Outdoor Event Recognition (ED)

    **Question Requirements:** Determine whether an event takes place indoors or outdoors.  
    
    **Question Description:** Identify the location setting of an event (indoor or outdoor) and provide evidence from the video, such as background elements or lighting conditions, to support your answer.
     
    \#\#\#\# Question Type 47: Main Characters in Events (ED)

    **Question Requirements:** Identify the main characters involved in an event. 
    
    **Question Description:** List the key characters participating in a specific event and describe their actions and interactions.
     
    \#\#\#\# Question Type 48: Event Dialogue Summary (ED)

    **Question Requirements:** Summarize the main dialogue during an event.  
    
    **Question Description:** Provide a brief overview of the key conversations that take place during an event and explain their significance to the plot or character development.
     
    \#\#\#\# Question Type 49: Event Conflict Identification (ED)

    **Question Requirements:** Identify the central conflict during an event.  
    
    **Question Description:** Describe the main source of tension or conflict within an event and discuss how it affects the characters and advances the plot.
     
    \#\#\#\# Question Type 50: Event Start and End Landmark Scenes (ED)

    **Question Requirements:** Specify the start and end times of an event using landmark scenes.  
    
    **Question Description:** Identify the landmark scenes that mark the beginning and end of an event, and explain their significance.
     
    \#\#\#\# Question Type 51: Cause and Effect of Events (ED)

    **Question Requirements:** Analyze the cause and effect relationships in events.
    
    **Question Description:** Describe the causes leading to an event and the effects that result from it, providing examples from the video to support your analysis.
     
    \#\#\#\# Question Type 52: Event Sequence Analysis (ED)

    **Question Requirements:** Analyze the chronological order of events in the video.
    
    **Question Description:** Explain the sequence of events, focusing on how they are ordered and their temporal relationships, providing examples from the video to support your explanation.

    \textbf{Video Captioning:}
    
    \#\#\#\# Question Type 53: Main Content Summary (VCap)

    **Question Requirements:** Summarize the main content of the video segment. 
    
    **Question Description:** Provide a concise overview of the primary events, actions, and themes present in the video segment.
     
    \#\#\#\# Question Type 54: Dialogue Summarization (VCap)

    **Question Requirements:** Summarize the primary dialogue exchanges in a scene. 
    
    **Question Description:** Capture the essence of the conversations between characters, highlighting the main points discussed and their significance to the plot.
     
    \#\#\#\# Question Type 55: Emotional Expression (VCap)

    **Question Requirements:** Identify and describe the emotional expressions of characters in the video.
    
    **Question Description:** Focus on the emotions conveyed by the characters through their facial expressions, body language, and dialogue, and explain their impact on the scene.
     
    \#\#\#\# Question Type 56: Weather Conditions (VCap)

    **Question Requirements:** Identify the weather conditions depicted in the video segment.  
    
    **Question Description:** Describe the weather, such as rain, sunshine, snow, or fog, and explain how it influences the mood or actions within the scene.
     
    \#\#\#\# Question Type 57: Vehicles (VCap)

    **Question Requirements:** Identify and describe the vehicles appearing in the video segment.  
    
    **Question Description:** Provide details about the types, colors, and movements of any vehicles seen in the video, and explain their relevance to the scene.
     
    \#\#\#\# Question Type 58: Animals (VCap)

    **Question Requirements:** Identify and describe any animals present in the video segment.  
    
    **Question Description:** Focus on the species, appearance, and behavior of the animals, and explain their role or significance within the scene.
     
   \#\#\#\# Question Type 59: Lighting (VCap)

    **Question Requirements:** Describe the lighting used in the video segment. 
    
    **Question Description:** Explain the lighting techniques, such as natural light, artificial light, shadows, and highlights, and discuss how they contribute to the scene’s atmosphere.
     
    \#\#\#\# Question Type 60: Background Elements (VCap)

    **Question Requirements:** Describe the background elements and their significance in the video segment.
    
    **Question Description:** Focus on the setting, props, and any notable items in the background, and discuss how they add context or depth to the scene.

    \textbf{Video Emotion Recognition:}
    
    \#\#\#\# Question Type 61: Primary Emotion Identification (VER)

    **Question Requirements:** Identify the primary emotion displayed by a character in a specific scene.  
    
    **Question Description:** Determine the main emotion that a character is expressing in the scene, supported by visual cues such as facial expressions, body language, and tone of voice.
     
    \#\#\#\# Question Type 62: Emotional Transition (VER)

    **Question Requirements:** Describe the emotional transition of a character throughout a scene.  
    
    **Question Description:** Explain how the character's emotion changes from the beginning to the end of the scene, including what events or interactions cause these changes.
     
   \#\#\#\# Question Type 63: Group Emotion Dynamics (VER)

    **Question Requirements:** Analyze the collective emotions of a group of characters in a scene.  
    
    **Question Description:** Identify the predominant emotion among a group of characters and describe how their emotions influence each other.
     
    \#\#\#\# Question Type 64: Conflict and Emotion (VER)

    **Question Requirements:** Examine the emotional responses during a conflict.  
    
    **Question Description:** Describe the emotions displayed by characters involved in a conflict and how these emotions escalate or resolve the conflict.
     
    \#\#\#\# Question Type 65: Subtle Emotional Cues (VER)

    **Question Requirements:** Identify subtle emotional cues that indicate a character's true feelings.  
    
    **Question Description:** Focus on less obvious signs of emotion, such as micro-expressions, tone of voice, and body language, to determine a character's underlying feelings.
     
    \#\#\#\# Question Type 66: Emotion and Dialogue (VER)

    **Question Requirements:** Assess the relationship between dialogue and emotion.  
    
    **Question Description:** Analyze how a character’s emotional state is reflected in their dialogue, including word choice, tone, and pacing.
 
    \#\#\#\# Question Type 67: Emotion and Environment (VER)

    **Question Requirements:** Describe how the environment influences a character's emotions.  
    
    **Question Description:** Examine how elements of the scene's setting, such as lighting, weather, and location, contribute to or reflect the characters' emotional states.
   
    \#\#\#\# Question Type 68: Emotional Resolution (VER)

    **Question Requirements:** Identify how a character's emotional arc resolves in a scene.  
    
    **Question Description:** Explain how the events of the scene lead to a resolution of the character's emotional journey, noting any shifts from negative to positive emotions or vice versa.
     
    \#\#\#\# Question Type 69: Contrasting Emotions (VER)

    **Question Requirements:** Compare and contrast the emotions of two characters in a scene.  
    
    **Question Description:** Analyze the differing emotional responses of two characters to the same event or interaction, highlighting the reasons for their contrasting feelings.
     
    \#\#\#\# Question Type 70: Non-Verbal Emotional Expression (VER)

    **Question Requirements:** Identify non-verbal cues used by characters to express emotions.  
    
    **Question Description:** Describe how characters use body language, facial expressions, and other non-verbal behaviors to convey their emotions, providing specific examples from the scene.
    
\end{scriptsize}
\end{quote}

\subsection{D.3 Q\&A Construction Prompt (For the Needle-in-a-Haystack task)}
\begin{quote}
\begin{scriptsize}
You are provided with multiple frames from a video. You must ask a highly detailed, needle-in-a-haystack type question based on the scenes depicted in the video frames, strictly following the [Question Format]. Each question must include 5 options (A, B, C, D, E), with only one correct option. The requirements for options and answer, are as follows:

1. All incorrect options and the correct option should be of similar length, neither too long nor too short.

2. Incorrect options should be clearly different from the correct option but closely related to the video content.

3. The correct option should not be answerable based solely on the question content; it should require information from the video content.

4. The correct option should be randomly distributed among the incorrect options.

5. The answer should include the letter of the correct option (A, B, C, D, E) and the reasoning process for how this answer was determined.

[Question Format] as follows:

**Question:** In the classroom scene, a girl with glasses and a red shirt is seen writing on a clipboard attached to a wooden door. Behind her, a boy in a white checkered shirt with a backpack approaches the same clipboard. The room in the background contains various musical instruments, including drums. What is the color of the pencil the girl is holding while writing on the clipboard?

- A. Blue

- B. Green

- C. Yellow

- D. Red

- E. Black

**Answer:** C. Yellow + It can be seen from the video that the girl is holding a yellow pencil while writing on the clipboard.

It is necessary to ensure the diversity of question types as much as possible.
    
\end{scriptsize}
\end{quote}

\newpage
\section{E More Experiment Results}
The following shows the accuracy of sub-tasks in eight video understanding tasks across all baseline MLLMs, except for the 'Needle in a Haystack' task. Since this task asks detailed questions about uniformly segmented video clips without specific sub-task types, its sub-task accuracy isn't separately reported.

\subsection{E.1 Accuracy for Video Classification (VC)}
\begin{figure*}[ht]
    \vskip -0.1in
    \begin{center}
    \centerline{\includegraphics[width=0.9\linewidth]{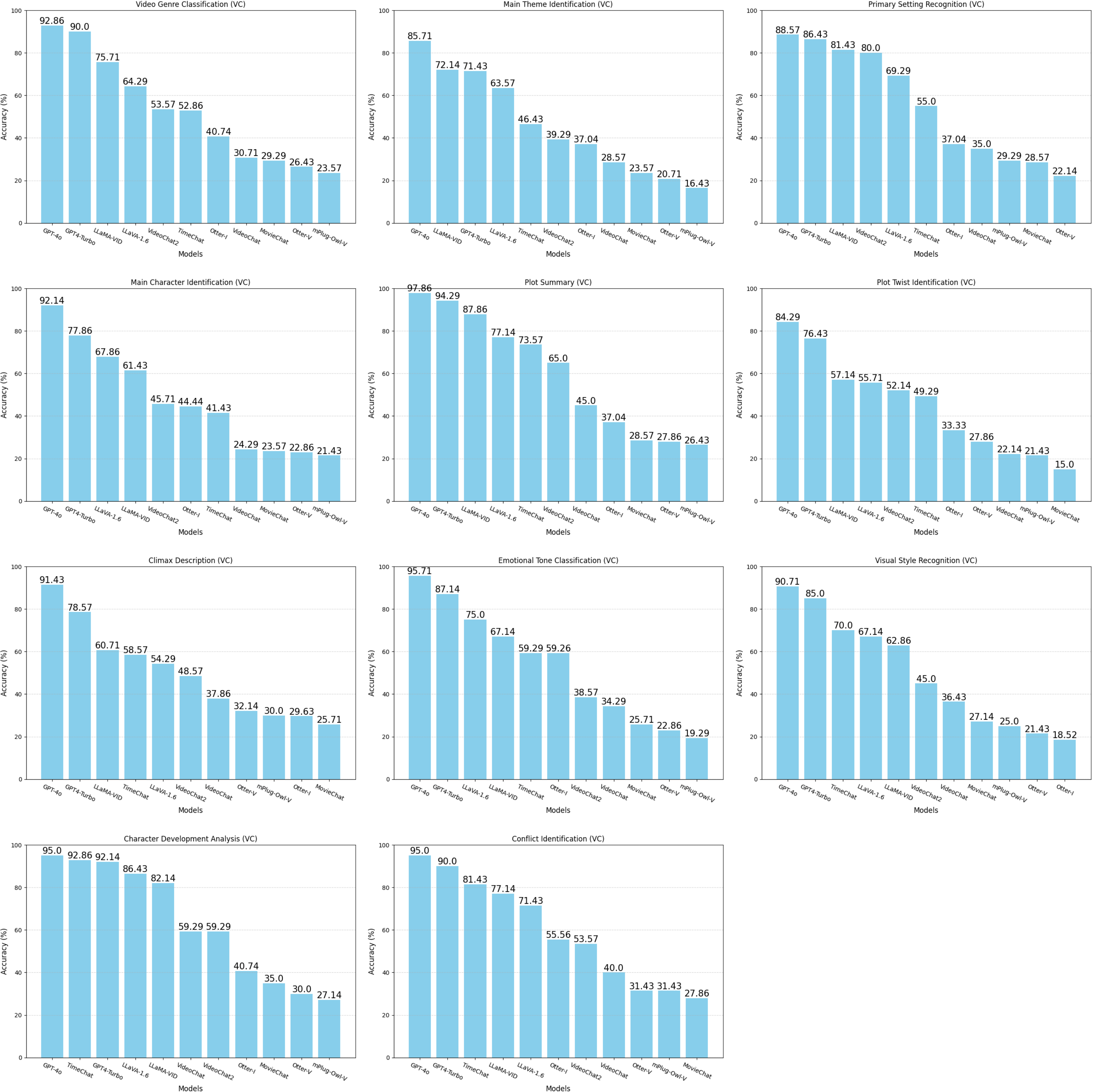}}
     \caption{The accuracy of sub-tasks in the VC task.}
     \label{acc_vc}
     \end{center}
\end{figure*}

\newpage
\subsection{E.2 Accuracy for Scene Recognition (SR)}
\begin{figure*}[ht]
    \vskip -0.1in
    \begin{center}
    \centerline{\includegraphics[width=0.9\linewidth]{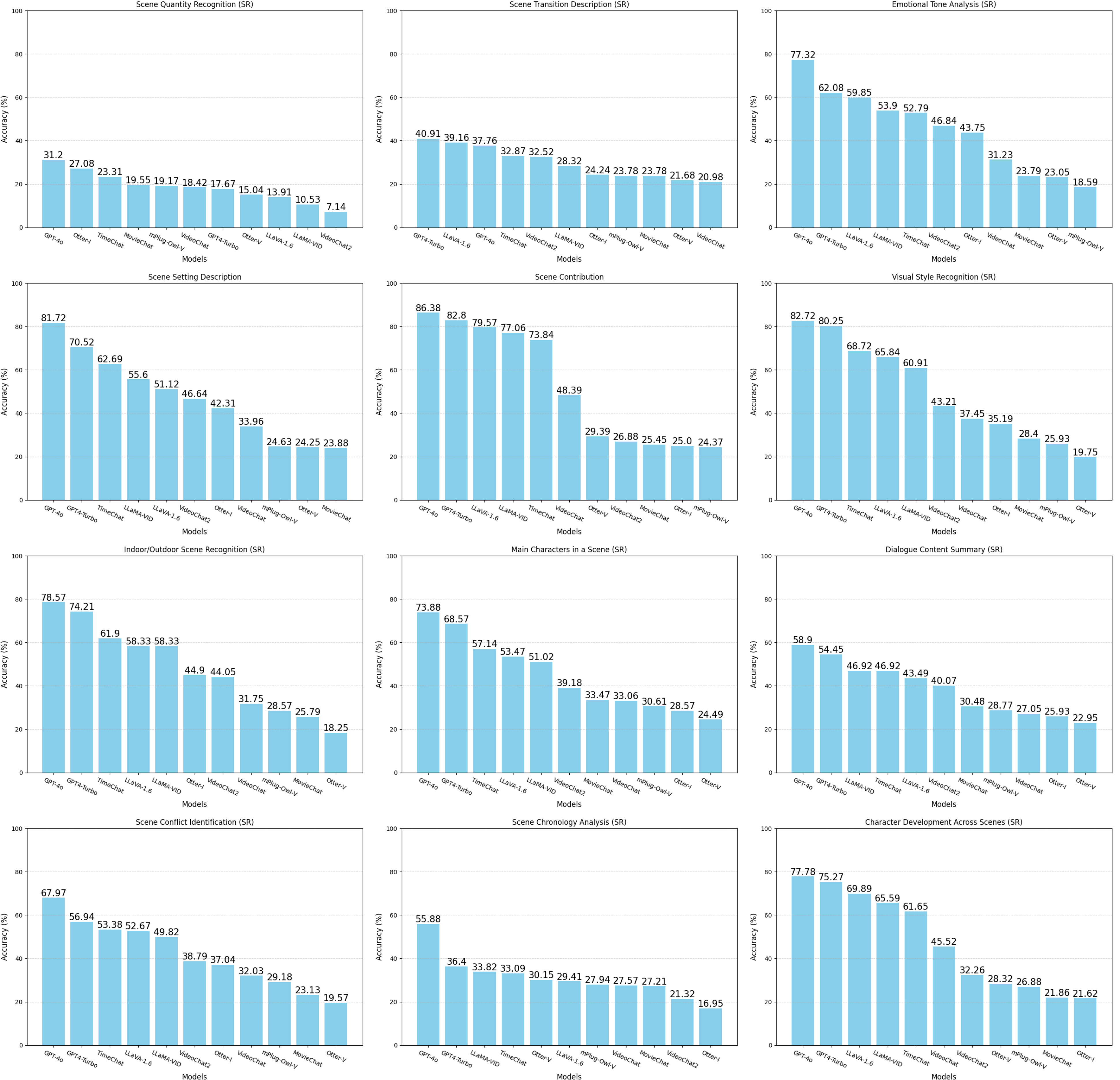}}
     \caption{The accuracy of sub-tasks in the SR task.}
     \label{acc_sr}
     \end{center}
\end{figure*}

\newpage
\subsection{E.3 Accuracy for Object Detection and Tracking (ODT)}
\begin{figure*}[ht]
    \vskip -0.1in
    \begin{center}
    \centerline{\includegraphics[width=0.9\linewidth]{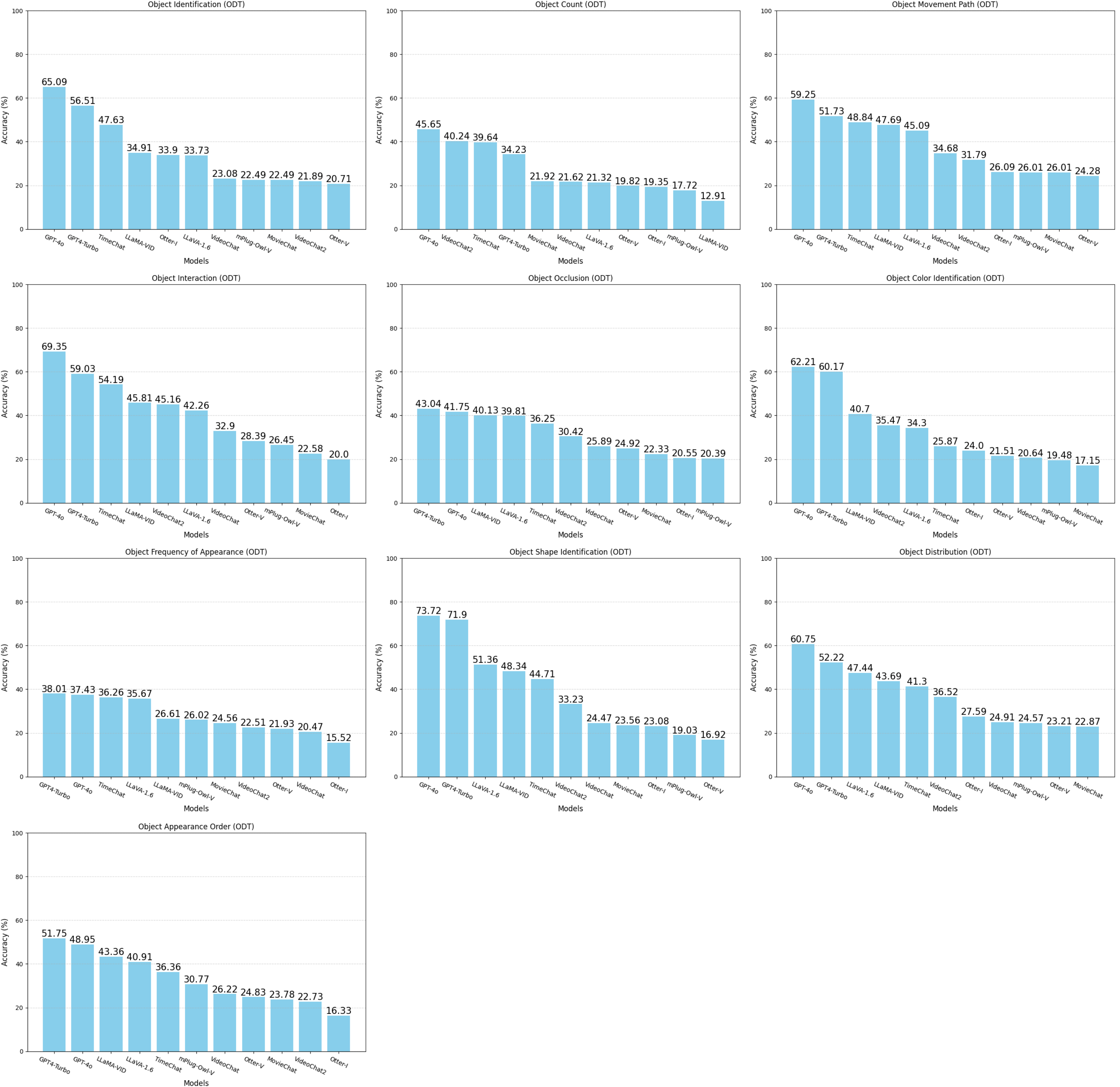}}
     \caption{The accuracy of sub-tasks in the ODT task.}
     \label{acc_odt}
     \end{center}
\end{figure*}

\newpage
\subsection{E.4 Accuracy for Action Recognition (AR)}
\begin{figure*}[ht]
    \vskip -0.1in
    \begin{center}
    \centerline{\includegraphics[width=0.9\linewidth]{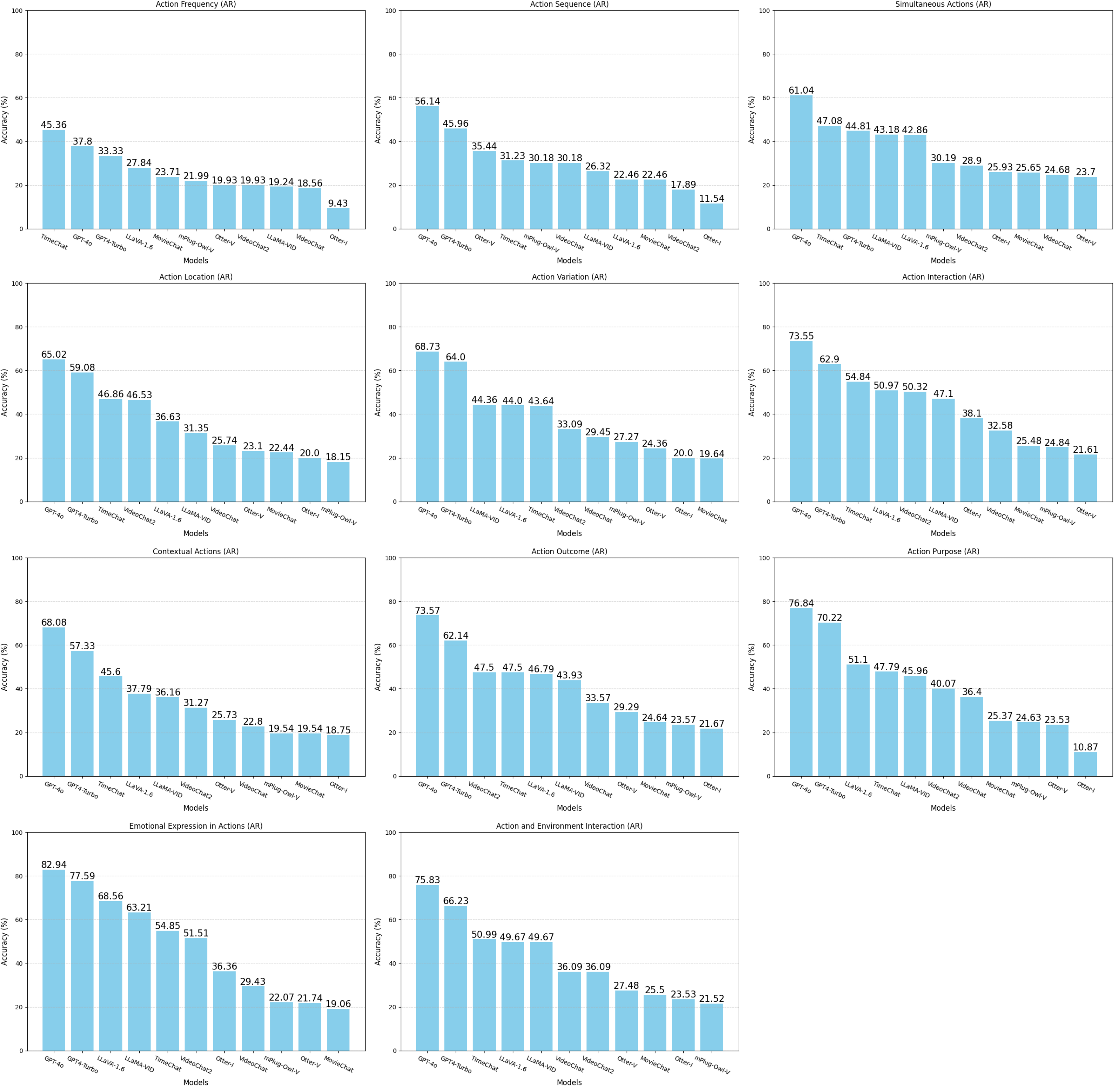}}
     \caption{The accuracy of sub-tasks in the AR task.}
     \label{acc_ar}
     \end{center}
\end{figure*}

\newpage
\subsection{E.5 Accuracy for Temporal Action Localization (TAL)}
\begin{figure*}[ht]
    \vskip -0.1in
    \begin{center}
    \centerline{\includegraphics[width=0.9\linewidth]{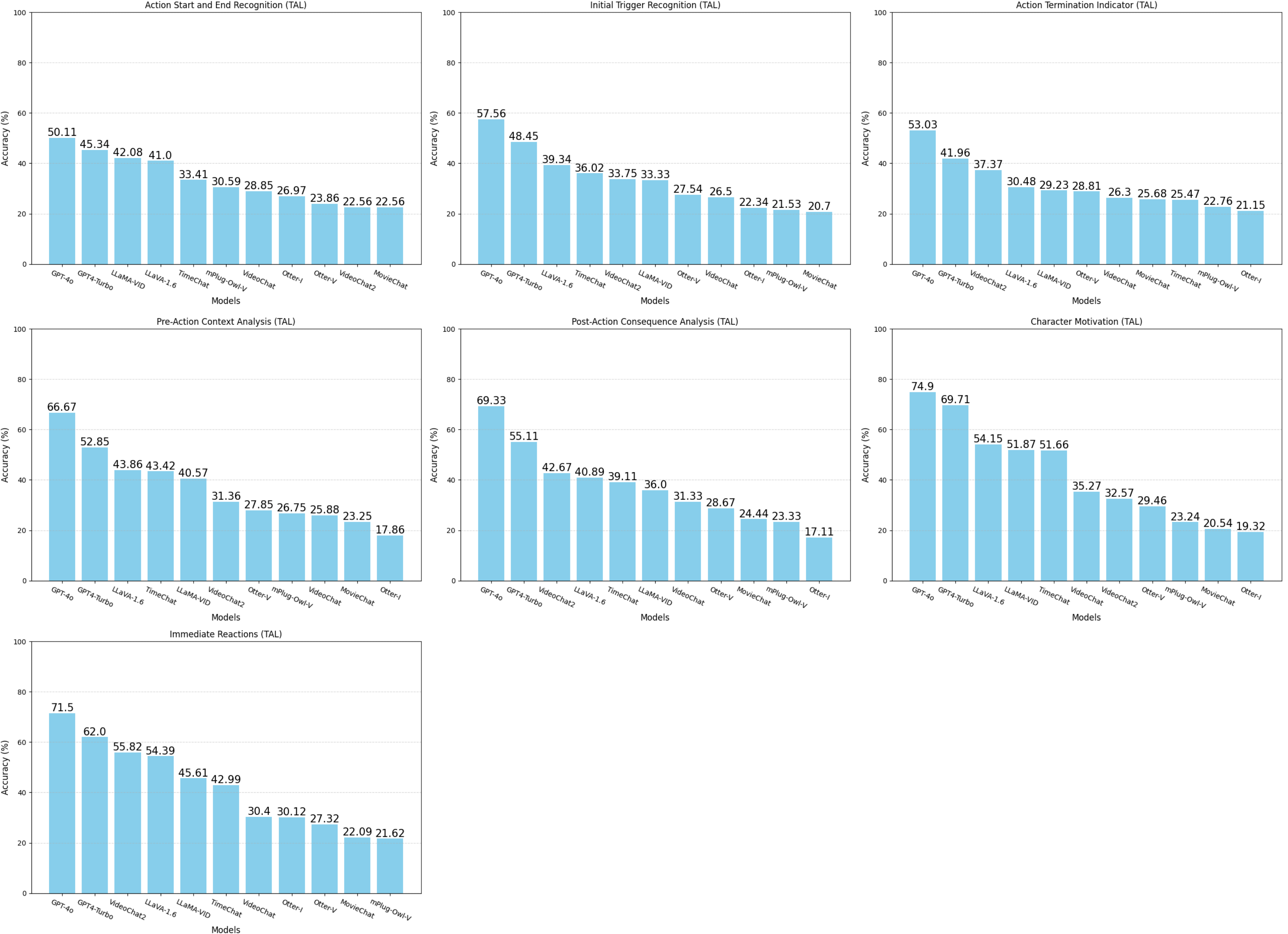}}
     \caption{The accuracy of sub-tasks in the TAL task.}
     \label{acc_tal}
     \end{center}
\end{figure*}

\newpage
\subsection{E.6 Accuracy for Event Detection  (ED)}
\begin{figure*}[ht]
    \vskip -0.1in
    \begin{center}
    \centerline{\includegraphics[width=0.9\linewidth]{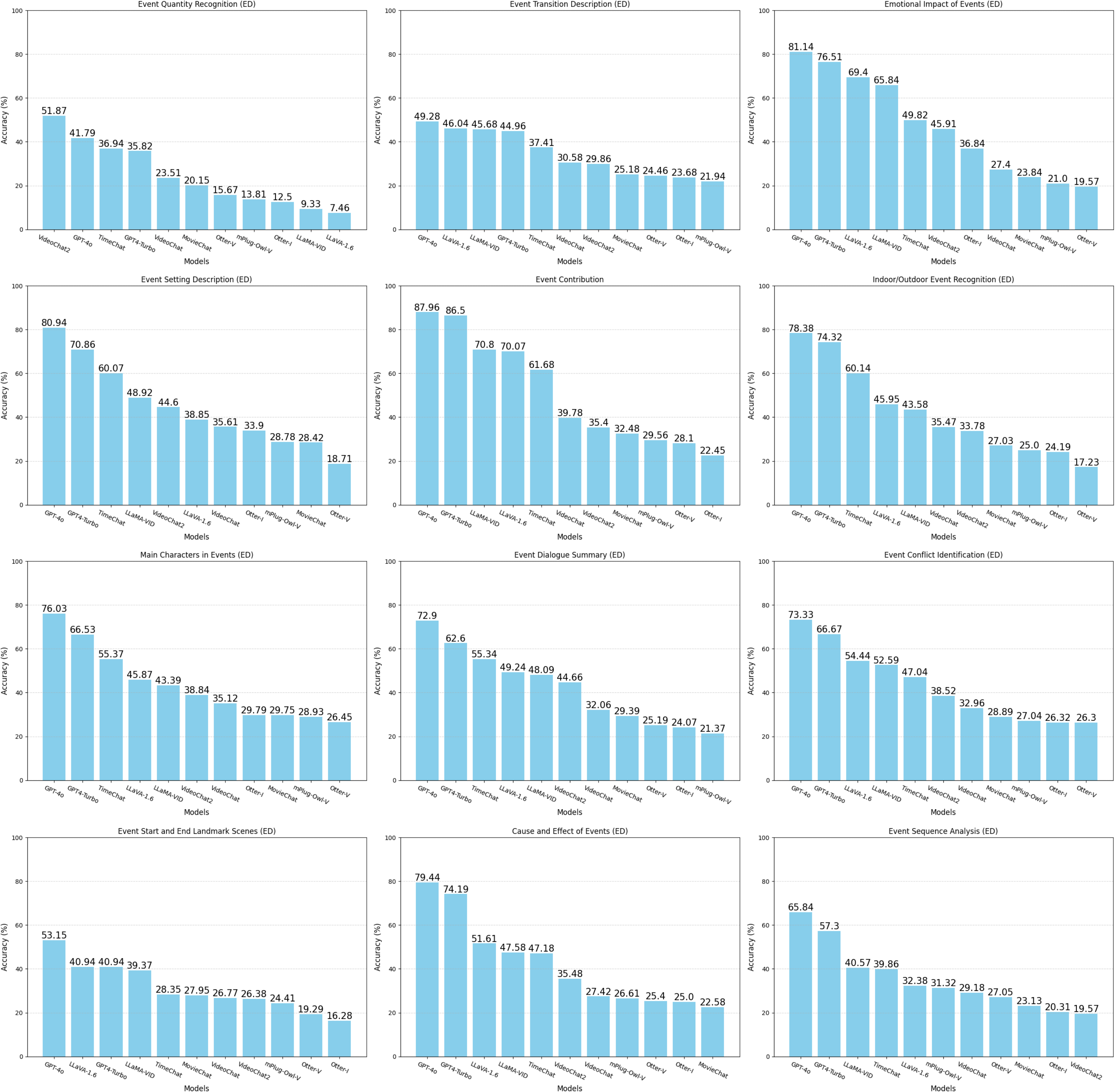}}
     \caption{The accuracy of sub-tasks in the ED task.}
     \label{acc_ed}
     \end{center}
\end{figure*}

\newpage
\subsection{E.7 Accuracy for Video Captioning  (VCap)}
\begin{figure*}[ht]
    \vskip -0.1in
    \begin{center}
    \centerline{\includegraphics[width=0.9\linewidth]{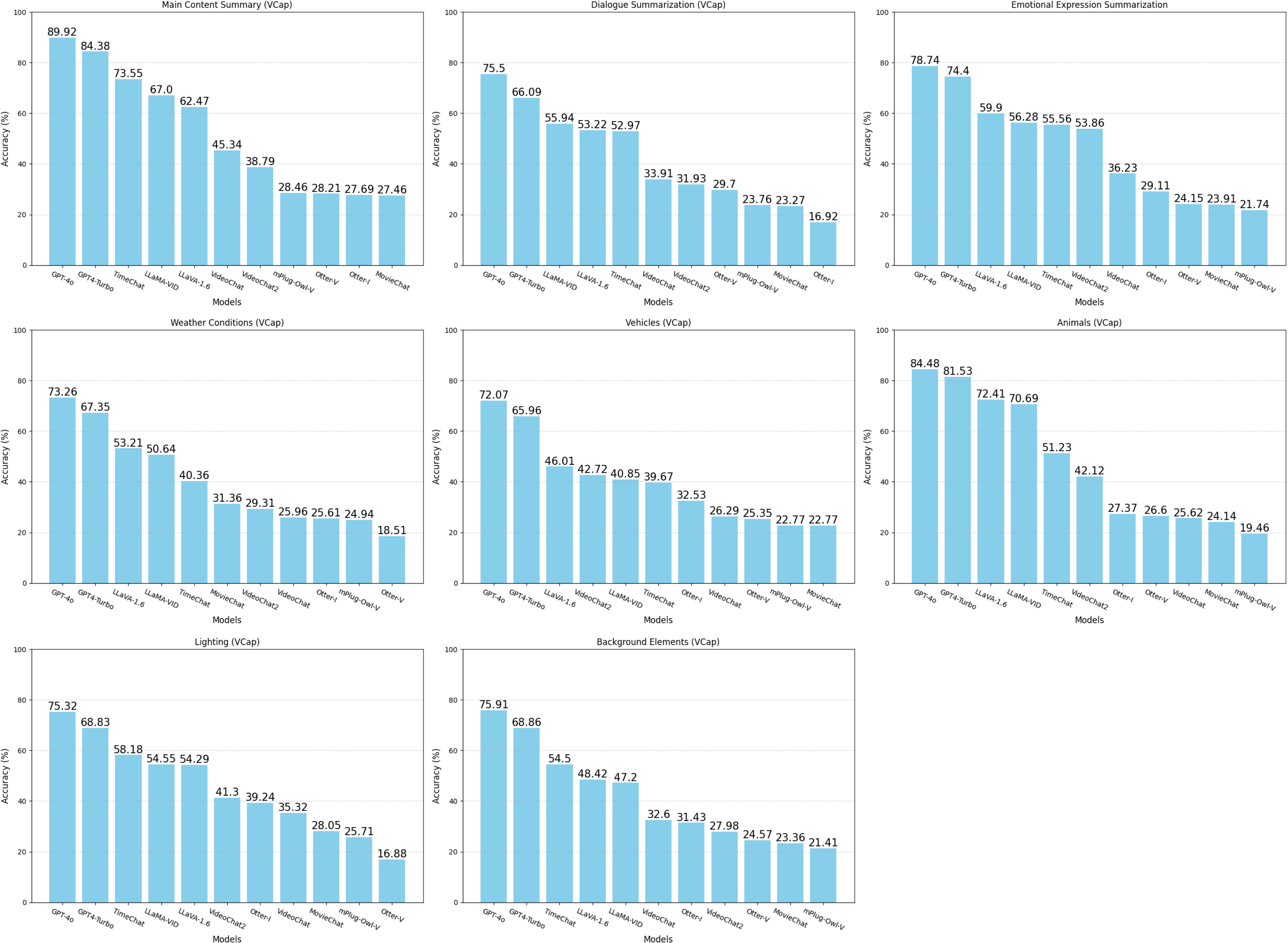}}
     \caption{The accuracy of sub-tasks in the VCap task.}
     \label{acc_vcap}
     \end{center}
\end{figure*}

\newpage
\subsection{E.8 Accuracy for Video Emotion Recognition (VER)}
\begin{figure*}[ht]
    \vskip -0.1in
    \begin{center}
    \centerline{\includegraphics[width=0.9\linewidth]{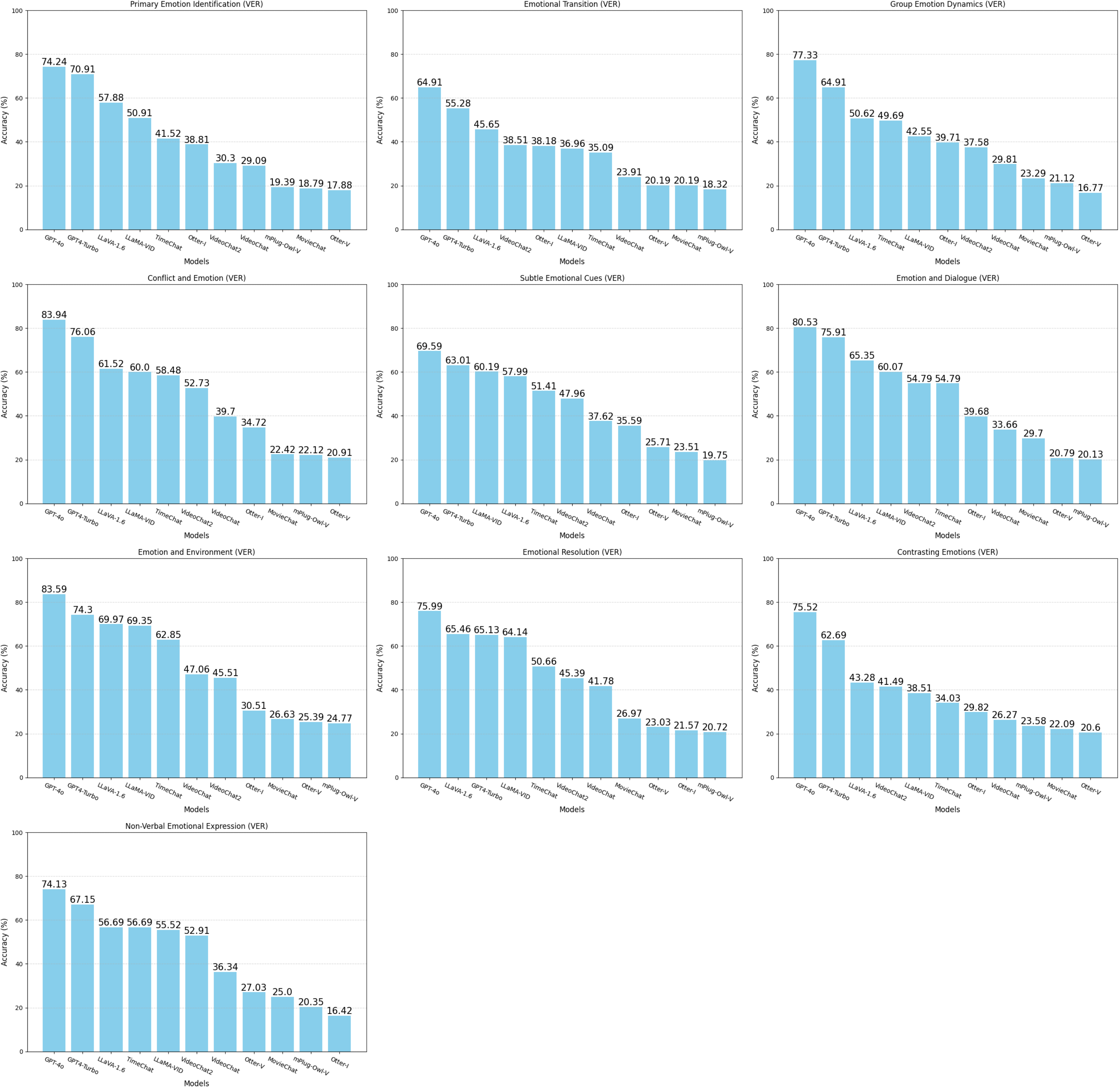}}
     \caption{The accuracy of sub-tasks in the VER task.}
     \label{acc_ver}
     \end{center}
\end{figure*}

\end{document}